\documentclass[10pt,journal,compsoc]{IEEEtran}
\ifCLASSOPTIONcompsoc
  \usepackage[nocompress]{cite}
\else
  \usepackage{cite}
\fi
\ifCLASSINFOpdf
\else
\fi
\usepackage{graphicx}
\usepackage{booktabs}
\usepackage{multirow}
\usepackage{rotating}
\usepackage{wrapfig}
\usepackage{amssymb}
\usepackage{amsmath}
\usepackage{caption}
\usepackage{subcaption}
\usepackage{threeparttable}
\usepackage{hyperref}
\usepackage{placeins}
\usepackage{afterpage}
\usepackage{amsfonts}
\usepackage{booktabs}
\usepackage{multirow}
\usepackage{makecell}
\usepackage{titlesec}
\usepackage{array}   
\usepackage{amssymb}
\usepackage[misc]{ifsym} 
\titleformat{\paragraph}
{\normalfont\normalsize\bfseries}{\theparagraph}{1em}{}
\titlespacing*{\paragraph}
{0pt}{3.25ex plus 1ex minus .2ex}{1.5ex plus .2ex}

\usepackage[table]{xcolor} 
\definecolor{rowcolor1}{RGB}{235,235,235} 
\definecolor{rowcolor2}{RGB}{250,250,250} 

\usepackage{tikz}

\usepackage[edges]{forest}
\begin{document}
%
\title{The Evolution of Dataset Distillation: Toward Scalable and Generalizable Solutions}
%
%
%
%

\author{Ping~Liu,~\IEEEmembership{Senior Member,~IEEE},
        and~Jiawei~Du\textsuperscript{~\Letter}
\IEEEcompsocitemizethanks{\IEEEcompsocthanksitem Ping Liu is with the Department of Computer Science and Engineering, University of Nevada, Reno, NV, 89512.\protect\\
E-mail: pino.pingliu@gmail.com
\IEEEcompsocthanksitem Jiawei Du is with Centre for Frontier AI Research (CFAR), and  Institute of High Performance Computing (IHPC), A*STAR, Singapore.\protect\\
E-mail: dujw@cfar.a-star.edu.sg
\IEEEcompsocthanksitem \textsuperscript{\Letter} denotes the corresponding author.
}
}

\IEEEtitleabstractindextext{%
\begin{abstract}
Dataset distillation, which condenses large-scale datasets into compact synthetic representations, has emerged as a critical solution for training modern deep learning models efficiently.
While prior surveys focus on developments before 2023, this work comprehensively reviews recent advances (2023–2025), emphasizing scalability to large-scale datasets such as ImageNet-1K and ImageNet-21K.
We categorize progress into a few key methodologies: trajectory matching, gradient matching, distribution matching, scalable generative approaches, and decoupling optimization mechanisms.
As a comprehensive examination of recent dataset distillation advances, this survey highlights breakthrough innovations: the SRe2L framework for efficient and effective condensation, soft label strategies that significantly enhance model accuracy, and lossless distillation techniques that maximize compression while maintaining performance.
Beyond these methodological advancements, we address critical challenges, including robustness against adversarial and backdoor attacks, effective handling of non-IID data distributions.
Additionally, we explore emerging applications in video and audio processing, multi-modal learning, medical imaging, and scientific computing, highlighting its domain versatility.
By offering extensive performance comparisons and actionable research directions, this survey equips researchers and practitioners with practical insights to advance efficient and generalizable dataset distillation, paving the way for future innovations.

\end{abstract}

}

\maketitle

\IEEEdisplaynontitleabstractindextext

%
\IEEEpeerreviewmaketitle

\IEEEraisesectionheading{\section{Introduction}\label{sec:introduction}}

\IEEEPARstart{T}he rapid advancement of large-scale deep learning models, such as Large Language Models (LLMs) \cite{minaee2024large_arxiv2024} and Vision-Language Models (VLMs) \cite{zhang2024vision_tpami2024}, has drastically increased the demand for large datasets to capture complex patterns and semantics. 
Although these models achieve state-of-the-art performance, their reliance on massive data poses significant challenges in terms of storage, computational cost, and energy efficiency, limiting accessibility and reproducibility. 
For instance, models like CLIP \cite{radford2021learning_icml2021} require over 400 million image-text pairs for pre-training, making dataset acquisition and processing prohibitively expensive. 
This raises concerns about the democratization of AI research, which only institutions with extensive computational resources can afford.

Dataset Distillation (DD) has emerged as a promising solution by condensing large datasets into compact synthetic representations that retain information critical for model training. 
First introduced by Wang et al. \cite{wang2018dataset_arxiv2018}, DD enables models to achieve performance comparable to full dataset training while significantly reducing storage and computational requirements. 
Beyond its practical benefits, recent theoretical breakthroughs in explaining the effectiveness of DD have further deepened the understanding of deep learning.
For instance, Yang et al. \cite{yang2024dataset_icml2024} demonstrated that distilled datasets primarily capture early training dynamics, closely resembling models that undergo early stopping on real data. 
By leveraging influence functions, they showed that individual distilled data points retain meaningful semantics unique to the target classes, offering critical insights into the fundamental mechanisms in dataset distillation.
These findings not only validate the effectiveness of DD but also pave the way for more principled distillation techniques.

Despite these advancements, dataset distillation faces several pressing challenges that hinder its large-scale deployment. 
One major challenge is scalability. 
Existing methods often fail to maintain performance when applied to large-scale datasets such as ImageNet-1K \cite{russakovsky2015imagenet} and ImageNet-21K \cite{ridnik2021imagenet_nips2021}. 
The nonlinear increase in computational overhead and the dramatic drop in performance on large-scale datasets pose significant challenges to scalability. 
Another critical limitation is cross-architecture generalization, where distilled datasets optimized for one model architecture often exhibit suboptimal performance on others. 
Furthermore, robustness concerns, including vulnerability to adversarial and backdoor attacks, must be addressed to ensure the security and reliability of distilled datasets. 
Beyond traditional image classification, DD is now being explored for other domains such as video, audio, multi-modal learning, and medical imaging, introducing new challenges that demand novel solutions. 
Given these evolving challenges and the rapid expansion of DD applications, a comprehensive survey is needed to consolidate recent innovations, identify key limitations, and outline future research directions.

As shown in Table \ref{tab:survey_comparison},  previous surveys \cite{geng2023survey_ijcai2023,lei2023comprehensive_tpami2023, yu2023dataset_tpami2023} focused primarily on early developments and small-scale datasets such as MNIST \cite{deng2012mnist_spm2012}, our work provides a timely and in-depth examination of the latest advances from 2023 to 2025. 
This survey differentiates itself by systematically addressing the scalability of DD methods to large datasets such as ImageNet-1K \cite{russakovsky2015imagenet} and ImageNet-21K \cite{ridnik2021imagenet_nips2021}, analyzing the effectiveness of cutting-edge techniques such as SRe2L series, soft-labeling strategies, and regularization-based approaches for improving performance at high IPC settings (e.g., IPC=200). 
We provide an in-depth comparison of different methodological paradigms, including trajectory matching, gradient matching, distribution matching, generative approaches, and decoupling optimization techniques, highlighting their respective strengths and limitations. 
Additionally, we explore underrepresented yet critical aspects of dataset distillation, such as adversarial and backdoor robustness against various attacks, self-supervised learning applications,  domain adaptations, and non Identical and Independent Distribution (IID).


\begin{table*}[t]
    \centering
 \caption{Comparison of recent dataset distillation surveys. Unlike previous surveys, which primarily focus on small datasets, low IPC settings, and supervised classification tasks, our work expands the scope to large-scale datasets, high IPC settings, generative approaches, decoupling-based solutions, and broader applications, such as self-supervised learning and federated learning.}
    \label{tab:survey_comparison}
    \begin{tabular}{lcccccc}
        \toprule
        \textbf{Survey}& \textbf{Coverage Period} & \textbf{Techniques Covered} & \textbf{Large-Scale } & \textbf{Large IPC} & \textbf{Beyond CLS} & \textbf{ SSL/DA/Non-IID}  \\
        \midrule
        Geng et al. \cite{geng2023survey_ijcai2023}  & 2019-2023 & GM, TM, DM  & $\times$   & $\times$  &  {\checkmark}  & $\times$  \\
        Lei et al. \cite{lei2023comprehensive_tpami2023}  & 2019-2023 & GM, TM, DM  &  $\times$    & $\times$  & $\times$ & $\times$    \\
        Yu et al. \cite{yu2023dataset_tpami2023}  & 2019-2023 & GM, TM, DM  & $\times$ & $\times$  & {\checkmark}    & $\times$  \\
        \textbf{Ours}  & 2023-2025 & GM, TM, DM, Generative, SRe2L, Softlabel & {\checkmark}  & {\checkmark}  &{\checkmark} & {\checkmark} \\
        \bottomrule
    \end{tabular}
\end{table*}

This survey is organized as follows. 
Section 2 introduces the background concepts and fundamental approaches in dataset distillation, covering techniques such as trajectory matching, gradient matching, and distribution matching. 
Section 3 highlights applications of dataset distillation across various domains and tasks, including temporal data, multi-modal learning, medical imaging, and scientific computing. 
Section 4 provides a comprehensive performance comparison of different methods on various benchmarks.
Finally, Section 5 discusses the current challenges in dataset distillation and outlines promising directions for future research.
Throughout our analysis, we aim to provide researchers and practitioners with actionable guidelines for advancing the field.

\begin{figure}[t]
    \centering
    \includegraphics[width=1.0\linewidth]{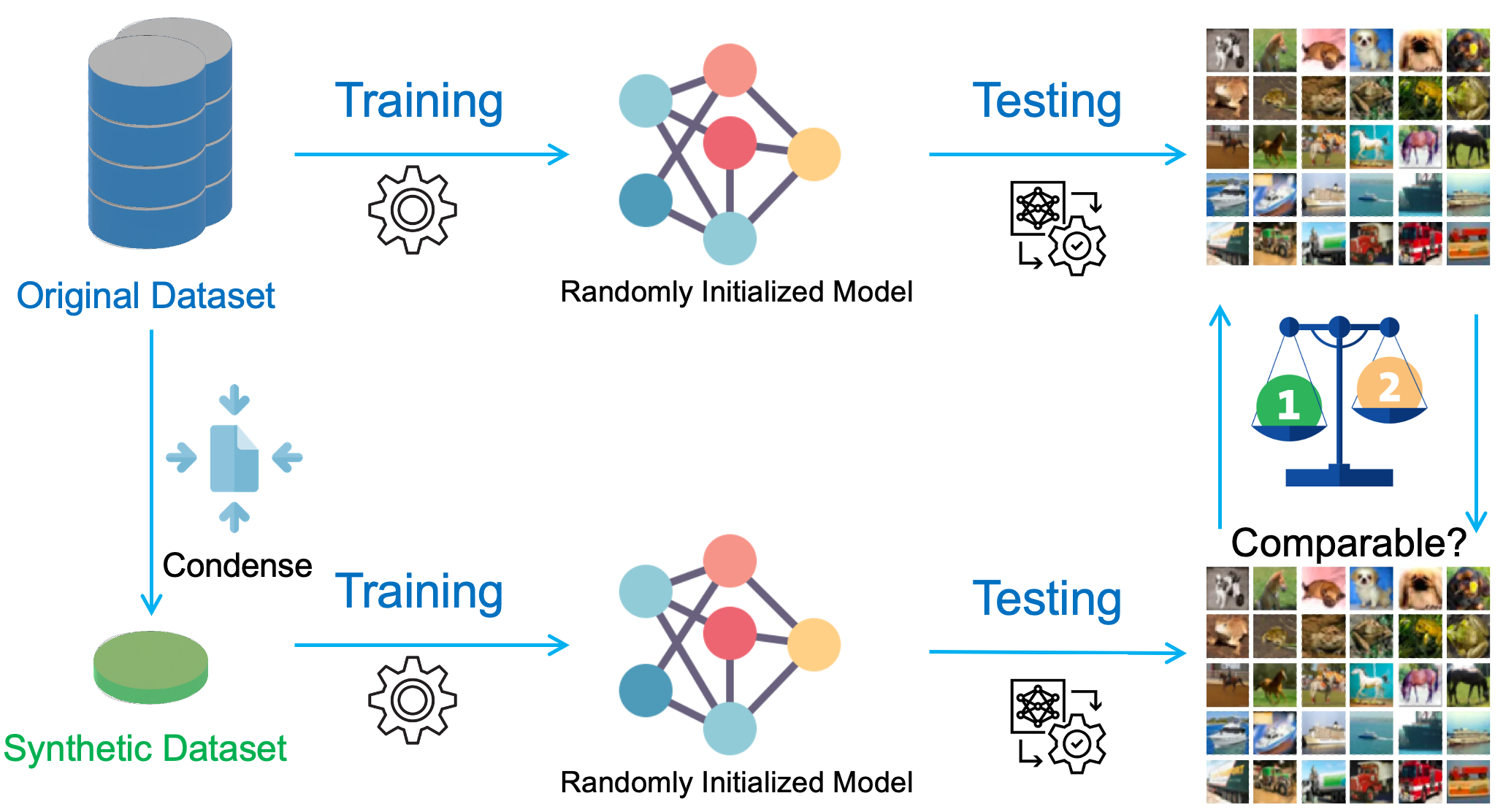}
\caption{Illustration of dataset distillation~\cite{wang2018dataset_arxiv2018}.  The original dataset is condensed into a synthetic dataset through a condensation process. Both the original and synthetic datasets are used to train randomly initialized models, and their performance is expected to be comparable. }
    \label{fig:gradientmatching}
\end{figure}

\section{Fundamental Dataset Distillation Methods}
\label{sec:background}
Dataset distillation, first introduced by Wang et al. \cite{wang2018dataset_arxiv2018}, seeks to condense a large training dataset into a significantly smaller, yet highly effective synthetic set.
The foundational formulation of dataset distillation is defined as:

\begin{equation}
\begin{split}
\mathcal{S} &= \arg\min_{\mathcal{S}} \mathbb{E}_{\theta^{(0)} \sim \Theta}[l(\mathcal{T};\theta^{(T)}_{\mathcal{S}})], \\
\theta^{(t)}_{\mathcal{S}} &= \theta^{(t-1)}_{\mathcal{S}}-\alpha\nabla_{\theta^{(t-1)}_{\mathcal{S}}} l(\mathcal{S}; \theta^{(t-1)}_{\mathcal{S}}),
\end{split}
\end{equation}
where $\mathcal{S}$ represents the synthetic training data to be learned, $\mathcal{T}$ denotes the original training dataset,  $\theta^{(t)}_{\mathcal{S}}$ represents model parameters at step t, $\theta^{(0)}$ is initial parameters, $\Theta$ is the distribution of initializations, $\alpha$ is learning rate, and $l(\cdot;\theta)$ represents the loss function.

As shown in Figure \ref{fig:gradientmatching}, the primary objective is to generate a condensed dataset that enables training new models with performance equivalent to those trained on the full original dataset.
Wang et al. employed backpropagation through time (BPTT) \cite{werbos1990backpropagation} as a core mechanism, emphasizing consistent model architectures across distillation and training phases to ensure alignment and stability.
This seminal work laid the foundation for extensive research aimed at enhancing the efficiency, generalization, and scalability of dataset distillation techniques.

However, traditional BPTT suffers from challenges like gradient variance, computational inefficiency, and difficulty capturing long-term dependencies. 
To address these limitations, subsequent research introduced innovative approaches.
Feng et al. \cite{feng2023embarrassingly_iclr2023} proposed Random Truncated BPTT (RaT-BPTT), which combines truncation and randomization to stabilize gradients and improve efficiency. 
This method effectively handles long-term dependencies, leading to significant performance gains in dataset distillation.

Building upon this,  Yu et al. \cite{yu2025teddy_eccv2024} developed the Teddy framework, which employs Taylor-approximated matching to simplify second-order optimization into a more computationally efficient first-order approach:
\begin{equation}
\begin{split}
l(\mathcal{T};\theta^{(T)}_{\mathcal{S}}) = l(\mathcal{T};\theta^{(T-1)}_{\mathcal{S}}-\alpha g_{\mathcal{S}}^{(T-1)}) \\
\approx l(\mathcal{T};\theta^{(T-1)}_{\mathcal{S}})-\alpha g_{\mathcal{T}}^{(T-1)}\cdot g_{\mathcal{S}}^{(T-1)} \\
\approx l(\mathcal{T};\theta^{(0)})-\alpha \sum_{t=0}^{T-1}g_{\mathcal{T}}^{(t)}\cdot g_{\mathcal{S}}^{(t)} .   
\end{split}
\end{equation}
By applying first-order Taylor expansion to unfold the unrolled computation graph, it decouples the bi-level optimization process. 
This transformation reduces computational complexity while avoiding the need to store and backpropagate through multiple training steps.

Building on the foundational concepts and innovations, following research has diversified into three primary matching-based approaches: distribution matching, which focuses on aligning the statistical properties of datasets; gradient matching, which seeks to replicate the training dynamics over short-term optimization steps by aligning gradients; 
and trajectory matching, which extends this idea to long-term consistency by aligning the entire optimization path of models trained on distilled data. 
While these methods share a common goal of preserving the training behavior of the original dataset, their differing scopes, such as short-term versus long-term alignment, address unique challenges in dataset distillation. 
The subsequent sections provide an in-depth exploration of these works, with emphasis on the advancements in the past two years.

\subsection{Matching Based Approaches}

\subsubsection{Gradient Matching} 
Dataset distillation through gradient matching represents one of the earliest and most fundamental approaches in matching-based methods. 
As shown in Figure \ref{fig:gradientmatching}, these methods aim to align the gradients of models trained on synthetic data with those trained on the original full dataset, ensuring that the distilled dataset guides the model towards similar optimization trajectories as the original dataset.

Zhao et al. \cite{zhao2020dataset_iclr2021} first formalized this approach through a gradient matching framework. 
The core idea is that a model   trained on distilled data should achieve both comparable generalization performance and parameter alignment with a model  trained on the original dataset.
The objective of gradient matching can be  defined as:
\begin{equation}
    \min_\mathcal{S} \mathbb{E}_{\theta_0\sim p(\theta_0)}[\sum_{t=0}^{T-1} D(\nabla_\theta\ell^\mathcal{S}(\theta_t), \nabla_\theta\ell^\mathcal{T}(\theta_t))],
\end{equation}
where $\theta$ is the model parameters, $\ell(\cdot)$ is the loss function,
$D(\cdot,\cdot)$ measures the distance between gradients, typically implemented using cosine similarity or L2 distance.
To further enhance the robustness and effectiveness of gradient matching,  Zhao et al. \cite{zhao2021dataset_icmlr2021} introduced Differentiable Siamese Augmentation (DSA).
This approach addresses the limited diversity of synthetic datasets by incorporating differentiable data augmentation into the matching process. 
DSA applies consistent transformations to both real and synthetic data while maintaining end-to-end gradient propagation, significantly improving the generalization capability.

\noindent \textbf{Stage Summary} However, while gradient matching effectively ensures step-wise optimization similarity, it faces several limitations.
Focusing on matching individual gradient steps may not capture long-term dependencies in training and can be sensitive to learning rate schedules and optimization hyperparameters.
Additionally, the computational cost of calculating and matching gradients at each step can be substantial, particularly for large models and datasets.
These limitations motivated researchers to explore methods that consider the entire training trajectory rather than individual gradient steps, leading to the development of trajectory matching approaches.

\subsubsection{Trajectory Matching}  
\begin{figure}[t]
    \centering
    \includegraphics[width=1.0\linewidth]{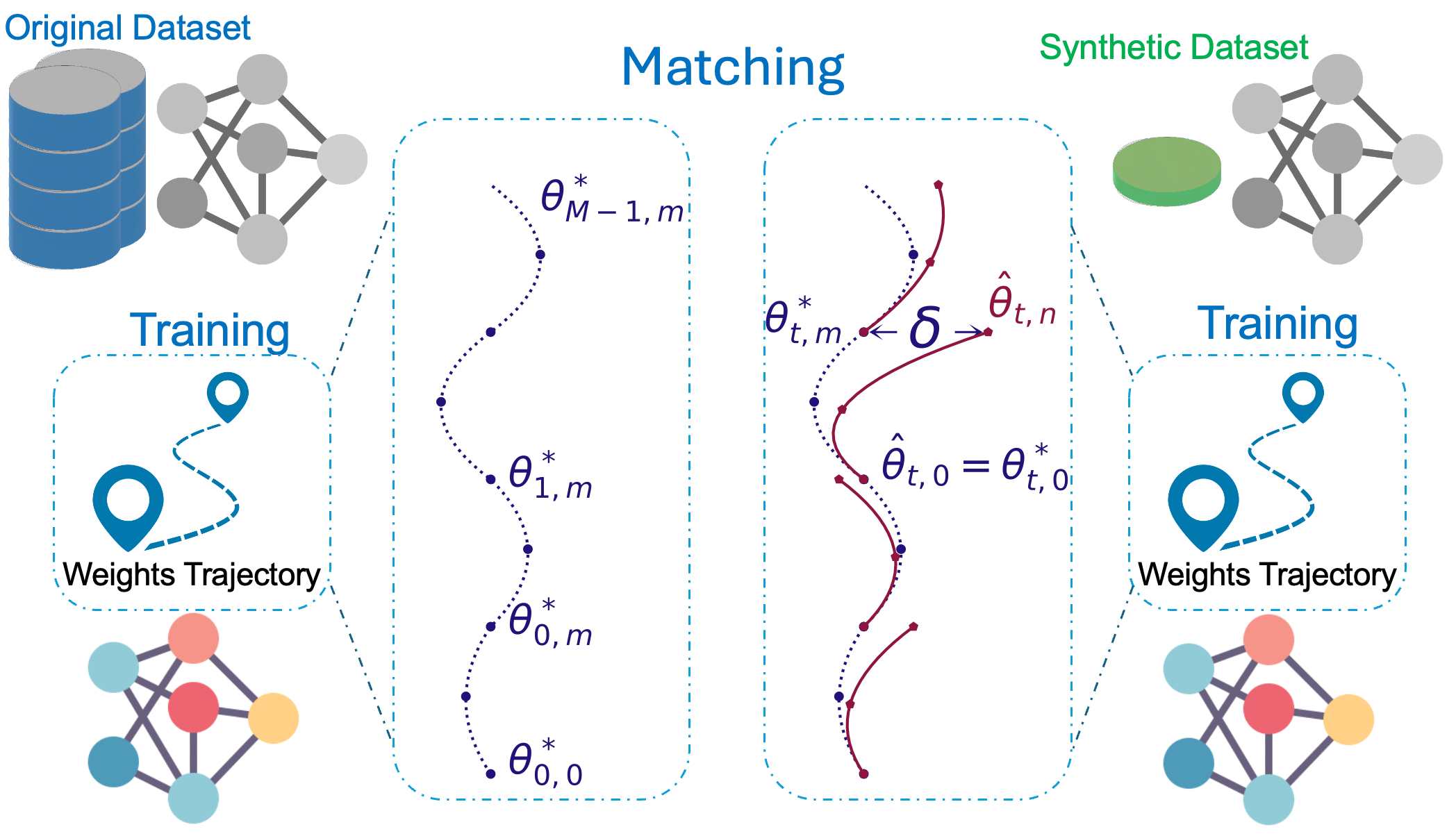}
    \caption{Overview of trajectory matching-based dataset distillation. The method aligns parameter trajectories between models trained on synthetic and real data. }
    \label{fig:trajectorymatching}
\end{figure}

Building upon the limitations of gradient matching, trajectory matching methods extend the optimization alignment from individual gradient steps to entire parameter trajectories during training. 
This holistic approach addresses both the long-term dependency issue and optimization instability inherent in gradient-based methods. 
As shown in Figure \ref{fig:trajectorymatching}, by aligning complete training trajectories between networks trained on synthetic and real data, this approach enables more robust and stable distillation.

Cazenavette et al. \cite{cazenavette2022dataset_cvpr2022} formalized this trajectory-level matching through their Matching Training Trajectories (MTT) method. 
While gradient matching focuses on step-wise gradient alignment, MTT uses expert trajectories from real datasets as benchmarks for the entire training process.
The formulation of MTT is defined as:
\begin{equation}
\mathcal{S}^* = \arg \min_{\mathcal{S}} \mathbb{E}_{\theta^{(0)} \sim \Theta} \sum_{t=0}^T \mathcal{D}(\theta^{(t)}_{\mathcal{S}}, \theta^{(t)}_{\mathcal{T}}),
\end{equation}
where $\theta^{(t)}_{\mathcal{S}}$ and $\theta^{(t)}_{\mathcal{T}}$ represent model parameters at step $t$ when trained on synthetic and real data respectively, and $\mathcal{D}(\cdot,\cdot)$ measures parameter-space distances rather than gradient differences.
The initialization distribution $\Theta$ and trajectory length $T$ are crucial hyperparameters that influence the robustness of the matching process \cite{du2023minimizing_cvpr2023}.
The development of trajectory matching methods has been driven by three primary challenges: \textit{trajectory stability}, \textit{parameter alignment}, and \textit{computational scalability}. 
Each subsequent advancement has contributed to addressing these challenges in complementary ways.

Ensuring stable and robust trajectories is crucial for effective dataset distillation.
While MTT effectively aligns trajectories, challenges like accumulated trajectory errors and sensitivity to weight perturbations persist. 
Du et al. \cite{du2023minimizing_cvpr2023} proposed Flat Trajectory Distillation (FTD), which reduces sensitivity by encouraging flat parameter trajectories through regularization. 
This approach improves robustness and compactness, particularly on larger datasets like CIFAR-100 \cite{krizhevsky2010convolutional}. Similarly, Shen et al. \cite{shen2023ast_arxiv2023} introduced Adaptive Smooth Trajectories (AST), incorporating gradient clipping, penalties, and representative sample initialization to stabilize training dynamics and mitigate trajectory mismatches.
Zhong et al. \cite{zhong2024towards_arxiv2024} addressed trajectory instability and slow  convergence with Matching Convexified Trajectories (MCT), which replaces oscillatory SGD paths with stable linear trajectories inspired by Neural Tangent Kernel (NTK) theory \cite{jacot2018neural_nips2018}. 
By requiring only two checkpoints, MCT reduces memory overhead while improving distillation stability.

Tackling parameter mismatches is also essential for distillation across heterogeneous datasets.
Li et al. \cite{li2024dataset_ieice2024} tackled the issue of parameter mismatches between teacher and student models caused by data heterogeneity.
Their method identifies mismatched parameters by evaluating the values of parameter magnitudes and prunes these selected parameters. 
The distilled dataset is then optimized over the remaining effective parameters, significantly enhancing distillation efficiency and performance.

To tackle scalability challenges with large-scale datasets, Cui et al. \cite{cui2023scaling_icml2023} developed Trajectory Matching with Soft Label Assignment (TESLA). 
TESLA refines the MTT loss function by decomposing gradient computations into sequential batches, eliminating the need to store full computational graphs in memory. 
Additionally, TESLA incorporates a soft label assignment strategy, leveraging pre-trained teacher models to generate soft labels for synthetic data, which significantly boosts performance when scaling to datasets with a large number of classes, such as ImageNet-1K.
Another innovation in this direction is the Automatic Training Trajectories (ATT) method introduced by Liu et al. \cite{liu2024dataset_eccv2024}. 
Unlike most trajectory matching approaches that use fixed trajectory lengths, ATT dynamically adjusts trajectory lengths during the distillation process. 
By employing a minimum distance strategy to select optimal matching trajectories, ATT effectively reduces mismatching errors and ensures precise alignment of synthetic datasets.
Kong et al. \cite{kong2025efficient_icassp2025} propose MTT-LSS, another trajectory matching-based method. 
Unlike previous methods emphasizing large-scale datasets, MTT-LSS primarily aims at reducing computational overhead and memory usage through low-rank space sampling. 
It achieves this by representing synthetic data with shared dimension mappers and low-dimensional basis vectors. 
Further evaluations on larger-scale datasets could help demonstrate its full potential regarding scalability.

\noindent \textbf{Stage Summary} Despite these advances, trajectory matching still faces important challenges. 
The high computational demands of tracking full parameter trajectories limit applications to complex architectures, while sensitivity to hyperparameters can affect robustness. 
Future research opportunities include developing adaptive hyperparameter selection mechanisms, designing more efficient algorithms for resource-constrained settings, and exploring hybrid approaches that combine the benefits of trajectory matching with other distillation techniques.

\subsubsection{Distribution Matching} 
Dataset distillation traditionally relied on gradient and trajectory matching,  suffering from two fundamental limitations: the computational complexity of bi-level optimization and the challenge of capturing long-term dependencies.
Distribution matching emerges as an alternative paradigm that addresses these limitations through direct alignment of feature distributions in selected embedding spaces.

\begin{figure}[t]
    \centering
    \includegraphics[width=1.0\linewidth]{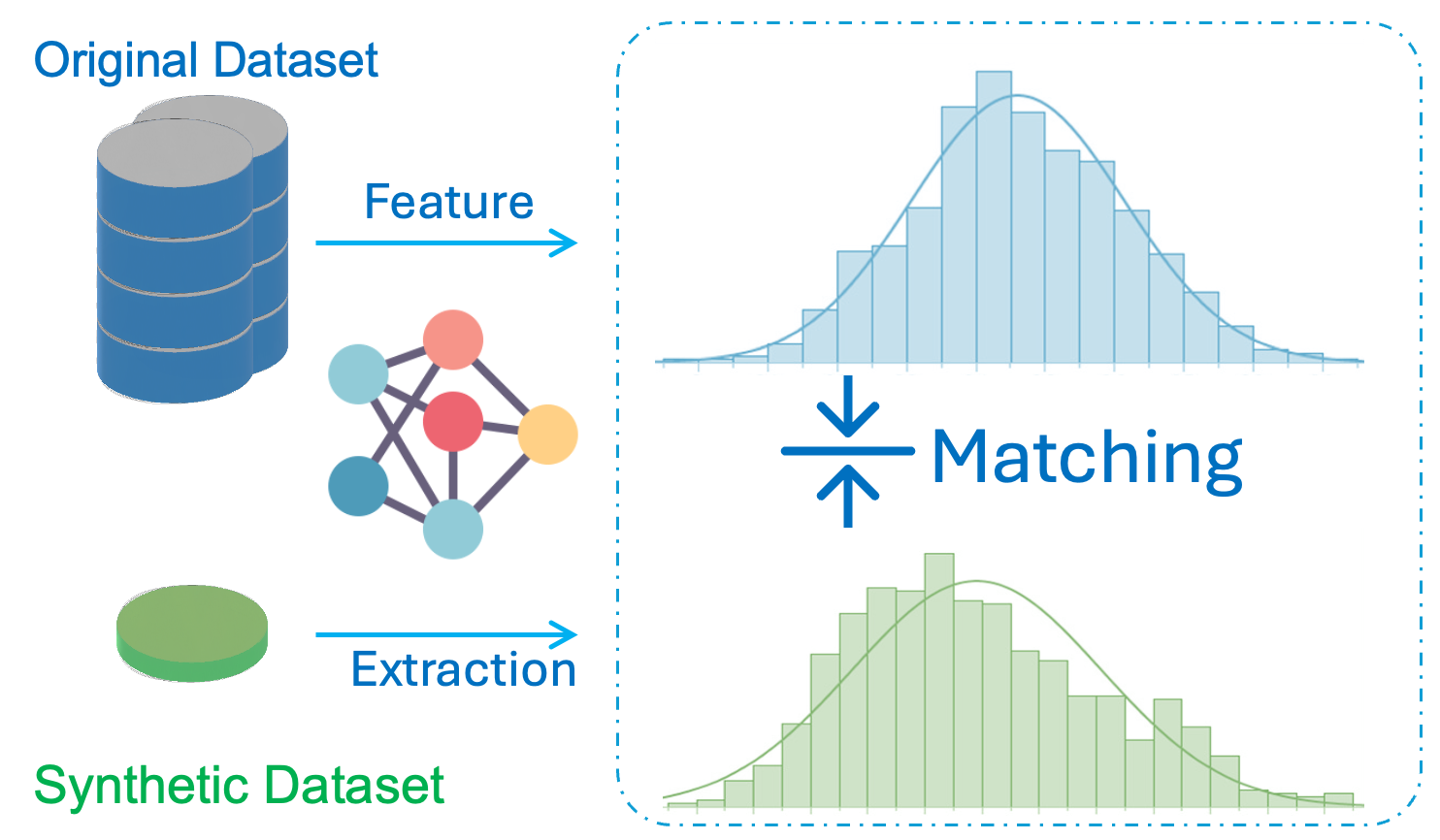}
    \caption{Overview of distribution matching in dataset distillation, where feature distributions are aligned to ensure the synthetic dataset effectively preserves the key characteristics of the original data.}
    \label{fig:distmatching}
\end{figure}

As shown in Figure \ref{fig:distmatching}, the core objective of distribution matching is to optimize synthetic data such that its distribution aligns with the original data distribution across multiple embedding spaces. 
This approach differs fundamentally from gradient or trajectory matching methods, offering improved computational efficiency and interpretability. 
The formal optimization problem is expressed as:
\begin{equation}
\mathcal{S}^* = \arg\min_{\mathcal{S}} \mathbb{E}_{\phi \sim \Phi}[\mathcal{D}(P_{\phi}(\mathcal{S}), P_{\phi}(\mathcal{T}))],
\end{equation}
where $P_{\phi}(\cdot)$ denotes the distribution of features extracted by network $\phi$, and $\mathcal{D}(\cdot,\cdot)$ measures the distance between distributions. 
This formulation, first introduced by Zhao et al. \cite{zhao2023dataset_wacv2023}, avoids the costly bi-level optimization process prevalent in earlier methods such as \cite{wang2018dataset_arxiv2018, zhao2020dataset_iclr2021}.
The following developments in distribution matching have focused on three major areas: \textit{feature refinement}, \textit{advanced alignment}, and \textit{higher-order matching}.

In the first area, the CAFE framework \cite{wang2022cafe_cvpr2022} introduced multi-scale feature preservation to address overfitting to dominant features.
This multi-scale approach significantly improves the discriminative power of synthetic datasets.
Zhao et al. \cite{zhao2023improved_cvpr2023} further refined this by introducing partitioning strategies and class-aware regularization to handle imbalanced feature distributions.
Zhang et al.’s \cite{Zhang2024_ijcai2024} DANCE framework addresses both distribution shifts and training inefficiencies.
By combining pseudo long-term distribution alignment with expert-driven distribution calibration, DANCE effectively aligns intra-class and inter-class distributions, achieving state-of-the-art performance.
Rahimi et al. \cite{Malakshan2024distilling_arxiv2024} took a different approach by decomposing distributions into content and style components, enabling more nuanced optimization of feature diversity.

In the second area, advanced alignment mechanisms have emerged as a critical development in dataset distillation, going beyond traditional statistical matching by incorporating attention-based strategies, trajectory constraints, and pseudo-labeling techniques. 
Sajedi et al. \cite{sajedi2023datadam_iccv2023} proposed DataDAM, which integrates feature distribution alignment with spatial attention matching. 
By leveraging attention maps aggregated from feature representations, DataDAM effectively emphasizes informative regions while reducing redundant background information, leading to more compact yet expressive synthetic datasets.

In the third area, recent approaches  utilized higher-order distribution matching, a crucial advancement beyond traditional mean distribution alignment.
Zhang et al. \cite{zhang2024m3d_aaai2024} pioneered embedding representations into a Reproducing Kernel Hilbert Space \cite{berlinet2011reproducing}, enabling alignment across multiple moments.
Wei et al. \cite{wei2024dataset_cvprw2024} complemented this with Latent Quantile Matching (LQM), aligning synthetic data with optimal quantiles determined via statistical tests.
From a geometric perspective, Liu et al. \cite{liu2023dataset_arxiv2023} leveraged the Wasserstein distance \cite{panaretos2019statistical}, capturing first and second-order statistics for distribution alignment.
Similarly, Deng et al. \cite{deng2024exploiting_cvpr2024} proposed class centralization and covariance matching, incorporating second-order information to tighten intra-class clustering while maintaining inter-class separation.
Wang et al. \cite{wang2025dataset_cvpr2025} propose a frequency-domain-based method that redefines distribution matching using the Neural Characteristic Function (NCF) and employs a minmax optimization strategy to dynamically learn an optimal discrepancy metric, ensuring a more precise alignment between distilled and real data distributions. 
The method first maximizes the discrepancy between real and distilled data in a learned frequency-space representation to establish the most discriminative distribution metric. 
Then, it minimizes this discrepancy by optimizing the distilled data to align with the real data under this learned metric. 
Unlike traditional methods that operate in pixel or feature space, this approach matches both phase  and amplitude information in the frequency domain, enhancing the realism and diversity of distilled data while significantly improving computational efficiency.

\noindent \textbf{Stage Summary} Collectively, these methods highlight the transition from simple mean-matching techniques to statistically and geometrically informed alignment strategies.
Notably, recent work by Li et al. introduced Hyperbolic Dataset Distillation (HDD) \cite{li2025hyperbolicdatasetdistillation_arxiv2025}, leveraging hyperbolic geometry to explicitly integrate hierarchical relationships into distribution matching, significantly improving representational effectiveness.
However, as shown in Table \ref{tab:performance}, distribution matching methods often lag behind gradient matching approaches in accuracy on large-scale datasets, revealing fundamental limitations in capturing dynamic training behaviors. 
To bridge this gap, future research could explore hybrid approaches that integrate distribution and gradient matching, develop adaptive higher-order constraints that evolve during optimization, and investigate causal relationships within feature distributions to gain deeper insights into the effectiveness of different matching strategies.

\subsubsection{Latent and Frequency Space Methods}
Traditional dataset distillation methods typically operate in the pixel space, directly generating synthetic data in the spatial domain. 
However, these methods often face significant computational and memory challenges, especially when handling high-dimensional datasets. 
To address these limitations, recent approaches have explored alternative spaces, such as latent and frequency domains, to achieve more efficient and scalable distillation.

Operating in latent spaces has emerged as a promising direction for efficient distillation due to its ability to reduce dimensionality while preserving essential information.
Duan et al. \cite{duan2023dataset_arxiv2023} proposed LatentDD, which transitions from the pixel space to the latent space using a pretrained autoencoder. 
By condensing datasets into compact latent representations, LatentDD substantially reduces computation and memory requirements, enabling efficient dataset distillation without compromising performance.
Building on this shift to latent spaces, George et al. \cite{cazenavette2023generalizing_cvpr2023} proposed GlaD, which leverages the latent spaces of generative models like StyleGAN-XL \cite{sauer2022stylegan_acmsiggraph}. 
By encoding high-dimensional images into low-dimensional feature vectors, GlaD not only reduces computational cost but also enhances generalization ability.
Further extending latent space methods, Li et al. proposed SLFD \cite{li2025datasetdistillationprobabilisticlatent_arxiv2025}, explicitly modeling spatial correlations and uncertainty within latent features using a low-rank multivariate normal distribution, effectively capturing structured uncertainty for improved dataset distillation.

Complementary to latent space approaches, frequency-domain methods offer another efficient avenue for dataset distillation.
Shin et al. \cite{shin2024frequency_nips2023} introduced Frequency Domain-based Dataset Distillation (FreD), a method that operates in the frequency domain rather than the spatial domain. 
FreD optimizes a selective subset of important frequency dimensions based on explained variance ratios, avoiding the need to process the full dimensionality of the spatial domain. 
This approach drastically lowers the memory budget per synthetic instance while preserving critical information. 
Building upon frequency-domain techniques, Yang et al. \cite{yang2024neural_eccv2024} introduced Neural Spectral Decomposition (NSD), leveraging the low-rank structure of datasets to decompose them into spectral tensors and kernel matrices, enhancing learning efficiency and integrating seamlessly with frameworks like MTT.
Moving beyond traditional representations, Shin et al. \cite{shin2025distilling_iclr2025} introduces Distilling Dataset into Neural Field (DDiF), a novel parameterization framework that compresses large datasets into synthetic neural fields under limited storage budgets, offering superior performance on various datasets.
To enhance the flexibility of frequency-based dataset distillation, Bo et al. \cite{bo2025understandingdatasetdistillationspectral_arxiv2025} proposed Curriculum Frequency Matching (CFM), which progressively refines filter parameters throughout the distillation process. 
Unlike static filtering methods, CFM dynamically transitions between low- and high-frequency components, improving adaptability and generalization.

\noindent\textbf{Stage Summary} These methods highlight the potential of alternative spaces to overcome the computational and scalability challenges inherent in pixel-based approaches. 
As latent and frequency space techniques continue to evolve, integrating their strengths or combining them with other methodologies could further enhance the efficiency and generalization of dataset distillation frameworks.

\subsubsection{Plug-and-Play Approaches}
Recent advancements in dataset distillation, such as \cite{du2024sequential_nips2024,cui2025optical_cvpr2025,belghazi2018mutual_icml2018,shang2024mim4dd_nips2023,zhong2024going_arxiv2024,son2024fyi_eccv2024,lu2023can_arxiv2023,zhong2024hierarchical_cvpr2025}, explored plug-and-play techniques that seamlessly integrate into existing methods, enhancing both optimization efficiency and generalization. 
Unlike approaches that introduce entirely new optimization paradigms, these methods serve as \textit{modular} enhancements, improving adaptability and scalability.
However, it is important to note that current plug-and-play techniques are primarily designed for matching-based methods, such as gradient matching and trajectory matching.

One major limitation in synthetic dataset optimization is treating the dataset as a single, unified entity during training, which can hinder the ability to capture training dynamics effectively. 
To tackle the uniform optimization limitation of synthetic datasets, Du et al. \cite{du2024sequential_nips2024} proposed Sequential Subset Matching (SeqMatch), a  strategy that partitions the dataset into smaller, manageable subsets. 
This approach addresses the inefficiencies associated with treating the entire dataset as a single unified whole, which often struggles to capture the dynamic nature of training processes. 
Instead of optimizing the dataset as an indivisible entity, SeqMatch divides it into  $K$  subsets, each sequentially optimized to capture knowledge relevant to specific stages of training.
The optimization process for each subset is expressed as:
\begin{footnotesize}
\begin{equation}
\hat{\mathcal{S}}_k = \arg\min_{\substack{\mathcal{S}_k \subset \mathbb{R}^d \times \mathcal{Y} \\ |\mathcal{S}_k| = \lfloor |\mathcal{S}| / K \rfloor}} 
\mathbb{E}_{\theta_0\sim P_{\theta_0}} \left[ \sum_{m=(k-1)n}^{kn} \mathcal{L}(\mathcal{S}_k \cup \mathcal{S}_{(k-1)}, \theta_m) \right],
\end{equation}
\end{footnotesize}
where K denotes the number of subsets, n is the number of iterations per subset.
Here,  $\mathcal{L}(.,.)$  represents the loss function used for model optimization,  $\mathcal{S}_{k-1}$  represents previously optimized subsets, and  $\mathcal{S}_k$  is the current subset being optimized. 
The cumulative incorporation of subsets ensures that knowledge captured in earlier stages is preserved and further refined in subsequent iterations.

Similarly, OPTICAL~\cite{cui2025optical_cvpr2025} addresses uniform optimization issues by introducing entropy-regularized Optimal Transport matching as a plug-and-play module.
Specifically, OPTICAL assigns different contribution weights from real to synthetic samples based on optimal transport matching of feature distances, allowing synthetic datasets to better preserve fine-grained geometric details. This leads to consistent improvements in generalization performance without requiring significant changes to existing gradient or trajectory matching frameworks.

Recently, mutual information maximization \cite{belghazi2018mutual_icml2018} has been introduced as a plug-and-play technique to enhance dataset distillation by preserving crucial feature representations. 
This approach seamlessly integrates with existing methods without requiring significant modifications to the core optimization framework, making it a flexible enhancement for various distillation techniques.
Shang et al. \cite{shang2024mim4dd_nips2023} introduced MIM4DD, reframing dataset distillation as a mutual information maximization problem. 
Using a contrastive learning framework \cite{khosla2020supervised_nips2020}, MIM4DD maximizes mutual information \cite{belghazi2018mutual_icml2018} between synthetic and real samples through positive pair attraction and negative pair repulsion within the representation space, while aligning samples from the same class and separating those from different classes.
Building on the perspective of mutual information maximization \cite{torkkola2003feature_jmlr2003}, Zhong et al. \cite{zhong2024going_arxiv2024} introduced class-aware conditional mutual information (CMI) as a plug-and-play regularization term. 
By concentrating synthetic datasets around class centers in pre-trained feature spaces, CMI improves generalization across architectures. 
This approach integrates seamlessly with existing techniques, offering consistent performance improvements across datasets.

Addressing biases in synthetic dataset generation is another critical focus.
Son et al. \cite{son2024fyi_eccv2024} tackled the bilateral equivalence phenomenon, where synthetic images symmetrically replicate discriminative parts, hindering the ability to distinguish subtle object details. 
To resolve this issue, they introduced FYI, a method that embeds horizontal flipping into the distillation process to break symmetry bias. 
This adjustment allows the synthetic datasets to capture richer and more nuanced object details, enhancing performance across multiple existing distillation techniques.

Leveraging external knowledge from Pre-Trained Models (PTMs) presents another promising avenue for improving dataset distillation.
Lu et al. \cite{lu2023can_arxiv2023} highlighted the untapped potential of PTMs, proposing Classification Loss of pre-trained Model and Contrastive Loss of pre-trained Model to transfer PTM knowledge into the distillation process.
These loss terms provide stable supervisory signals, leveraging diverse PTM architectures, parameters, and domain expertise. 
The study demonstrated the effectiveness of PTMs in enhancing cross-architecture generalization, even when using less-optimally trained models.

\noindent\textbf{Stage Summary} These plug-and-play approaches demonstrate the potential to seamlessly integrate advanced techniques into existing dataset distillation workflows. 
By improving efficiency, addressing biases, and leveraging external knowledge, these methods pave the way for scalable and generalizable distillation frameworks that can adapt to diverse datasets and architectures.

\subsection{Scalable Dataset Distillation Methods}
As the demand for large-scale and high-resolution datasets continues to grow, dataset distillation methods face mounting scalability challenges. 
To address these, recent advancements have introduced innovative frameworks that balance efficiency and performance, extending the applicability of dataset distillation to larger datasets. 
Key developments include the exploration of generative models such as GANs \cite{goodfellow2014generative} and Diffusion Models \cite{croitoru2023diffusion}, decoupling strategies of the SRe2L series \cite{yin2024squeeze_nips2024}, and the soft label techniques \cite{qin2024label_arxiv2024}. 
The following subsubsections delve into these cutting-edge methods, emphasizing their contributions and the challenges they address in achieving scalability.
\subsubsection{Dataset Distillation via Generative Models}
Recent advancements in dataset distillation have increasingly leveraged generative models to enhance scalability, efficiency, and generalization. 
Unlike traditional methods that rely on discriminative matching processes to create synthetic data, these approaches utilize generative mechanisms to synthesize high-quality, diverse datasets.

\paragraph{GAN-based Methods}
Early efforts in this direction focused on GANs. 
Zhao et al. \cite{zhao2022synthesizing_arxiv2022} introduced Informative Training GAN (IT-GAN), challenging the traditional focus on visual realism in GAN-generated data. 
Instead, IT-GAN prioritizes informative training samples by freezing a pre-trained GAN and optimizing latent vectors using condensation loss and diversity regularization.
In IT-GAN,  condensation loss is to  minimize discrepancies between real and synthetic data, and  regularization loss  promotes diversity in the generated data.

Expanding this direction, Li et al. \cite{li2024generative_cvpr2024} emphasized the importance of balancing global structures and local details in dataset distillation. 
Their method combines logits matching for capturing global structures with feature matching for refining local details, leveraging a model pool of diverse architectures to improve cross-architecture generalization.
Similarly, Wang et al. \cite{wang2023dim_arxiv2023} introduced Distill into Model (DiM), a novel approach that distills datasets into a generative adversarial network instead of directly producing synthetic images. 
DiM dynamically synthesizes diverse training samples from random noise, enhancing scalability for large architectures and eliminating the need for repeated re-distillation.
Building on these advancements, Zhang et al. \cite{zhang2023dataset_arxiv2023} proposed a codebook-generator framework, which condenses datasets into a compact codebook and generator.
This approach incorporates intra-class diversity and inter-class discrimination losses, enabling efficient synthesis of class-specific images and scalability to large datasets such as ImageNet-1K.
These methods collectively demonstrate the versatility and scalability of generative models across  architectures and  datasets in dataset distillation.

\paragraph{Diffusion-based Methods}
Diffusion models, which are renowned for their superior image quality compared to GANs, have emerged as a powerful alternative. 
Gu et al. \cite{gu2024efficient_cvpr2024} pioneered a diffusion-based framework leveraging a ‘minimax’ criterion to balance representativeness and diversity in surrogate datasets, ensuring both coverage and variability of training data. 
Similarly, Su et al. \cite{su2024d_cvpr2024} introduced Dataset Distillation via Disentangled Diffusion Model (D4M), which combines latent diffusion models with prototype learning. 
This approach enhances distillation efficiency and cross-architecture generalization while employing a novel training-time matching strategy. 
By eliminating architecture-specific optimization during image synthesis, D4M achieves high-resolution image generation, representing a substantial leap forward in dataset distillation methodologies.
To address the misalignment between generation and evaluation objectives, CaO2~\cite{wang2025_iccv2025} introduces a two-stage framework that first selects class-consistent samples using a lightweight classifier, then refines their latent codes to improve conditional alignment.
This design effectively mitigates misalignment between generation and evaluation objectives, leading to improved distillation quality across diverse settings.
To enhance semantic control and mitigate object inconsistency in generated images, Zou et al.~\cite{zou2025_iccv2025} proposed a vision-language distillation framework that introduces category-wise text prototypes alongside image prototypes to guide diffusion-based image generation.
By leveraging large language models to produce cluster-specific descriptions and integrating them into the sampling process, their method improves semantic coherence, reduces object omissions, and achieves superior accuracy across various benchmarks.

Exploring the capabilities of DMs further, Abbasi et al. \cite{abbasi2024one_arxiv2024} proposed Dataset Distillation using Diffusion Models (D3M), which compresses entire image categories into textual prompts using techniques like textual inversion and patch selection. 
This approach enables efficient dataset compression while maintaining high performance on large-scale benchmarks. 
In a complementary approach, Abbasi et al. \cite{2024arXiv241204668A_arxiv2024} combined coreset selection \cite{sun2024diversity_cvpr2024} with latent diffusion models, enhancing patch diversity and realism, achieving significant improvements in large-scale benchmarks.

Recently, latent space techniques have offered innovative ways to optimize storage efficiency, scalability, and adaptability in this direction.
Moser et al. \cite{moser2024latent_arxiv2024} introduced Latent Dataset Distillation using Diffusion Models (LD3M), which extends the distillation process into the latent space. 
By leveraging a pre-trained latent diffusion model, LD3M enhances gradient propagation through a modified diffusion process that integrates initial latent states, striking an optimal balance between speed and accuracy:

\begin{equation}
    z_{t-1} \leftarrow \left[\left(1-\frac{t}{T}\right) \cdot \mu_{\theta}(c, z_t, \gamma_t) + \frac{t}{T} \cdot z_T\right] + \sigma_t^2\varepsilon_t,
\end{equation}
where $T$ denotes the total number of diffusion steps, $t$ represents the current timestep in the backward process, $\mu_{\theta}$ is the parameterized mean predictor network with learnable parameters $\theta$, and ${z}_T$ is the initial noisy state. 
The linear interpolation coefficients $(1-\frac{t}{T})$ and $\frac{t}{T}$ control the balance between the mean prediction and initial state, providing enhanced gradient flow during training without compromising the generation quality.
This framework demonstrates superior performance for high-resolution tasks while integrating seamlessly with existing algorithms. 
Recently, Zhao et al. proposed D³HR~\cite{zhao2025taming_icml2025}, a diffusion-based dataset distillation framework leveraging DDIM inversion to map complex latent distributions into a structured Gaussian space for better distribution matching. 
Additionally, their novel group sampling strategy selects representative subsets based on statistical similarity, significantly improving accuracy, stability, and cross-architecture generalization compared to prior methods.
Taking a different approach to latent space optimization, Qin et al. \cite{qin2024distributional_arxiv2024} proposed Distributional Dataset Distillation (D3) and applied it into the federated learning setting \cite{10571602_tpami2024}. 
In their method, D3 represents classes as Gaussian distributions in latent space, optimizing the parameters to generate images using a decoder network. 
To guide the distilled data generation process, in \cite{chen2025igd_iclr2025}, Chen et al. introduced Influence-Guided Diffusion (IGD) that uses a latent diffusion model based on a Diffusion Transformer \cite{peebles2023scalable_iccv2023} to generate synthetic training datasets. 
They designed an influence guidance function to steer diffusions towards generating training-effective data and a deviation guidance function to enhance diversity, enabling the method to significantly improve performance across various datasets, particularly on high-resolution ImageNet-1K benchmarks.
Chan et al. \cite{chan2025mgd_icml2025} proposed MGD3, a training-free dataset distillation framework that leverages pre-trained diffusion models and injects semantic mode guidance directly into the sampling process to generate diverse and representative samples—without any model parameter updates.

\noindent\textbf{Stage Summary} While current methods have demonstrated promising results in improving scalability and efficiency through various generative approaches, several critical research directions deserve further investigation. 
First, with the rapid advancement of generative models, particularly the emergence of flow matching \cite{lipman2022flow_arxiv2022,chen2024flow_iclr2024} and other innovative approaches demonstrating superior effectiveness and efficiency, incorporating these cutting-edge generative techniques into dataset distillation presents a promising avenue for exploration. 
Second, current dataset distillation methods face significant challenges in memory usage and computational efficiency when handling high-resolution images. 
In this context, distillation approaches based on generative models, might offer potential solutions for high-resolution scenarios, inspiring further research into optimizing algorithmic architectures to enhance memory efficiency while maintaining distillation quality. 
Finally, the integration of generative approaches with traditional matching-based methods warrants thorough investigation.

\subsubsection{Decoupling Optimization: SRe2L Series}
\begin{figure}[t]
    \centering
\includegraphics[width=1.0\linewidth]{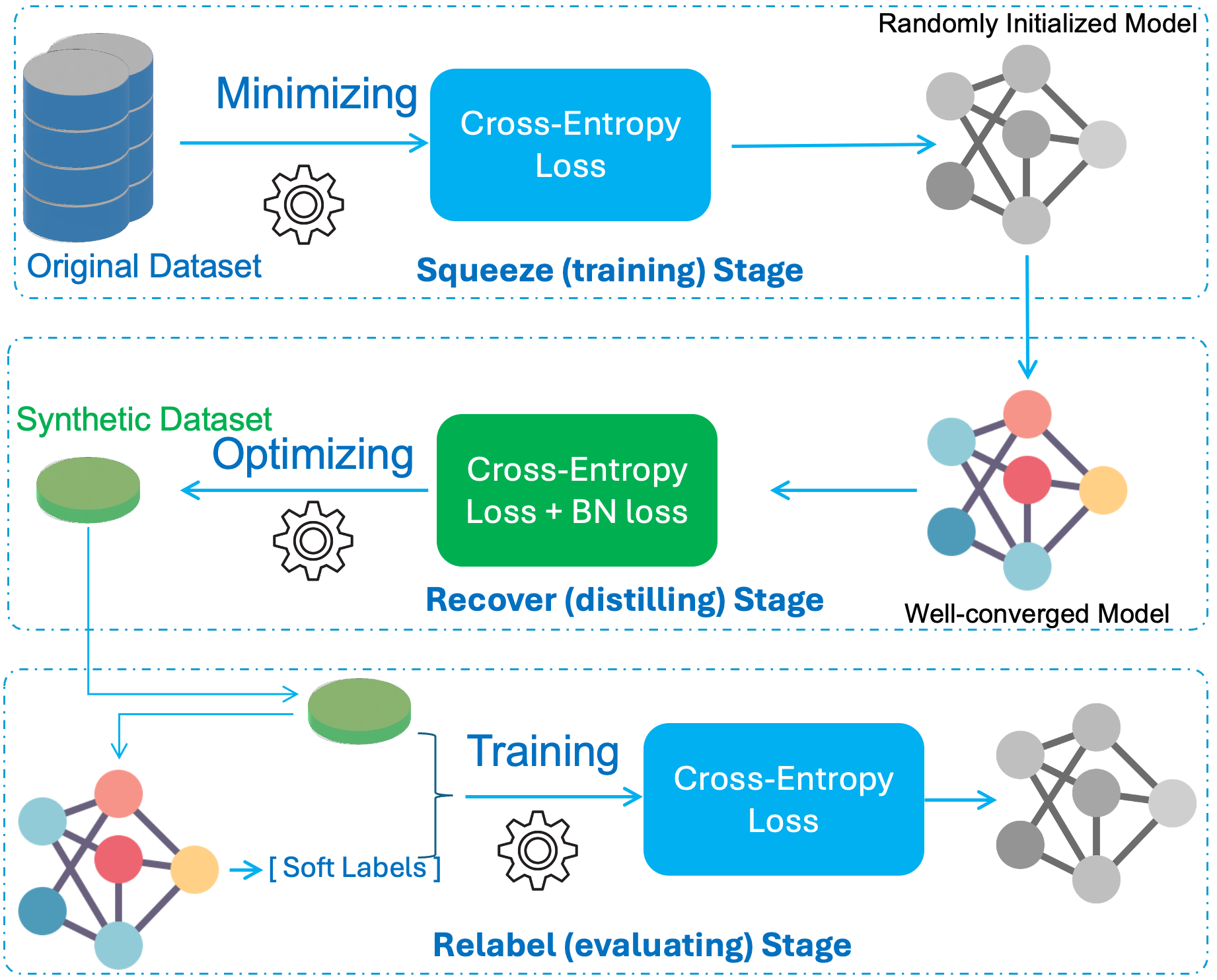}
    \caption{Overview of the three-stage SRe2L work \cite{yin2024squeeze_nips2024}. The Squeeze Stage trains a randomly initialized model on the original dataset by minimizing cross-entropy loss. The Recover Stage refines the synthetic dataset by optimizing a combined cross-entropy and batch normalization (BN) loss with a well-converged model. Finally, the Relabel Stage evaluates performance by training another model on the synthetic dataset with soft labels, using cross-entropy loss.}
    \label{fig:sre2l}
\end{figure}

The high computational costs associated with bilevel optimization have long hindered the scalability of dataset distillation methods, particularly when applied to large-scale datasets or high-resolution images. 
To address these challenges, Yin et al. \cite{yin2024squeeze_nips2024} proposed the Squeeze, Recover, and Relabel (SRe2L) framework, a three-stage process that decouples dataset condensation into manageable steps. 
As shown in Figure \ref{fig:sre2l}, the Squeeze stage extracts essential information by training a model on the original dataset, followed by the Recover stage, which generates synthetic data by aligning batch normalization (BN) statistics \cite{cai2024batch_cvpr2024} with the trained model. 
Finally, the Relabel stage assigns soft labels to the synthetic data using the model. 
This approach achieved promising performance on large-scale datasets such as ImageNet-1K at  resolutions of 224$\times$224 pixels.

Building upon this foundational work, Yin et al. introduced Curriculum Data Augmentation (CDA) \cite{yin2023dataset_arxiv2023}, which employs a curriculum learning approach \cite{jiang2015self_aaai2015} during the data synthesis process. 
By progressively increasing the difficulty of image crops through dynamic adjustments of RandomResizedCrop parameters, CDA captures global structures early in training and refines local details in later stages. 
This method set a milestone by successfully distilling the entire ImageNet-21K dataset \cite{ridnik2021imagenet_nips2021} at standard resolutions, marking an huge advance in large-scale dataset distillation.

Subsequent works systematically addressed the inherent limitations of SRe2L in scalability, generalization, and computational efficiency.
Zhou et al. \cite{zhou2024self_arxiv2024} introduced the SC-DD framework, which overcomes SRe2L’s challenges with larger models by enhancing BN statistics through self-supervised pretraining \cite{zong2024self_pami2024}. 
By enriching these statistics and employing linear probing during relabeling, SC-DD significantly improves data recovery and scalability for large datasets like ImageNet-1K. 
Shao et al. \cite{shao2024generalized_cvpr2024} extended this progress with the Generalized Various Backbone and Statistical Matching (G-VBSM) method, which addresses SRe2L’s reliance on single backbone models or specific statistics by introducing data densification, generalized statistical alignment, and multi-backbone consistency. 
G-VBSM achieved significant improvement across datasets and architectures, particularly in highly compressed settings. 
Further refinement came with the Elucidating Dataset Condensation (EDC) \cite{shao2024elucidating_nips2024}, which incorporated real-image initialization, category-aware matching, and flatness regularization. 
More recently, Cui et al. \cite{cui2025datasetdistillationcommitteevoting_arxiv2025} introduced Committee Voting for Dataset Distillation (CV-DD), synthesizing high-quality distilled datasets by integrating insights from multiple pre-trained models. 
CV-DD introduces a Prior Performance Guided voting strategy that assigns greater weights to better-performing pre-trained models in the voting process, and  proposes Batch-Specific Soft Labeling, which generates more accurate soft labels by recalculating BN statistics for each synthetic data batch.

\noindent \textbf{Stage Summary} Recent works have shifted from single-model approaches to leveraging model diversity, as seen in G-VBSM’s multi-backbone architecture and CV-DD’s committee voting mechanism.
These methods emphasize improving batch normalization statistics, with SC-DD enhancing BN through pre-training and CV-DD introducing batch-specific BN recalculation, leading to more robust, generalizable condensed datasets.

\subsubsection{Soft Labels} 
In the SRe2L series and beyond, soft labels \cite{yin2024squeeze_nips2024} have emerged as a critical component for the significant performance improvement, encoding rich semantic information that surpasses the limitations of traditional one-hot labels. 
These nuanced labels enhance diversity, generalization, and scalability, addressing core challenges in large-scale dataset condensation.
Highlighting the centrality of soft labels in efficient learning, Qin et al. \cite{qin2024label_arxiv2024} proposed to explore the role of soft labels in dataset distillation. 
Through extensive experiments, they demonstrated that early-stopped expert models generate soft labels encoding structured semantic information crucial for data-efficient learning.

Subsequent research has advanced both understanding and  implementations on soft labels.
Sun et al. \cite{sun2024diversity_cvpr2024} proposed the Realistic, Diverse, and Efficient Dataset Distillation (RDED) method, which employs Fast Knowledge Distillation \cite{shen2022fast_eccv2022} to generate region-level soft labels. 
By aggregating diverse features and representations, RDED moves beyond traditional patch selection methods, capturing more intricate semantic information. 
Extending these principles, Hu et al. \cite{hu2025focusddrealworldsceneinfusion_arxiv2025} introduced Focused Dataset Distillation (FocusDD), leveraging pre-trained Vision Transformers (ViT) \cite{50650_iclr2021} to identify critical image patches. 
Their approach combines these key regions with downsampled background information to create high-quality distilled datasets. 
Notably, FocusDD extends dataset distillation beyond classification tasks to object detection, showcasing the versatility of soft label approaches. 
Zhong et al. \cite{zhong2024efficientdatasetdistillationdiffusiondriven_arxiv2024} further advanced the field by incorporating DMs for soft label generation and synthetic image construction. 
Unlike earlier random region selection techniques, their method employs differential loss computation guided by text prompts to identify class-relevant regions, offering a powerful alternative on patch selection.
Those advances show a clear progression from simple patch selection to sophisticated region processing using Vision Transformers and Diffusion Models, significantly enhancing the quality of distilled datasets.

Addressing practical implementation challenges, Shang et al. \cite{shang2024gift_arxiv2024} proposed the GIFT framework. 
This approach combines a label refinement module—merging smoothed hard labels with teacher-generated soft labels—and a mutual information-based loss function. 
GIFT significantly enhances the effectiveness of distilled datasets while maintaining minimal computational overhead, demonstrating its adaptability across diverse scenarios.
Xiao et al. \cite{xiao2024large_nips2024} tackled the substantial storage demands of soft labels in large-scale distillation, which can exceed 30 times the size of the distilled dataset.
Their solution leverages class-wise batching and supervision during image synthesis to enhance within-class diversity, enabling effective soft label compression through simple random pruning.
This strategy dramatically reduces storage requirements while improving performance on large-scale datasets, making soft label-based methods more practical for real-world applications.

\noindent \textbf{Stage Summary} This progression in soft label methodologies represents a fundamental advancement in dataset distillation. 
Future research directions include exploring efficient and adaptive data and/or patch generation, investigating multi-modal soft label fusion. 
These developments will be crucial for extending the applicability of dataset distillation to more challenging datasets with large-scale and high resolution, as well as emerging domains such as self-supervised learning and large foundation models.

\subsection{Efficiency and Effectiveness in Dataset Distillation}
Recent innovations have introduced strategies that reduce computational overhead, optimize resource utilization, and enhance representational diversity, effectively addressing challenges such as high memory requirements, overfitting, and scalability. 
These advancements significantly improve performance on large-scale datasets.

\subsubsection{Selective Dataset Distillation}
Recent advances highlight the necessity of identifying and optimizing key components that contribute most to model performance.
We present a unified theoretical framework for selective approaches across multiple dimensions, demonstrating how strategic selection and weighting of elements, from low-level features to high-level representations, enhance distillation efficiency.
Formally, let $\mathcal{D} = {(x_i, y_i)}_{i=1}^N$ denote the original dataset and $\mathcal{S} = {(\tilde{x}_j, \tilde{y}_j)}_{j=1}^M$ represent the synthetic dataset, where $M \ll N$.
The selective distillation objective can be formulated as:
\begin{equation}
\min_{\mathcal{S}} \mathbb{E}_{\theta_0} \sum_{t=0}^{T-1} \omega_t \sum_{d \in \mathcal{D}} \omega_d \mathcal{L}_d(\mathcal{S}, \mathcal{T})
\end{equation}
where $\omega_t$ represents selection weights across different timesteps of the training trajectory, $\omega_d$ denotes selection weights for different dimensions (e.g., pixel, color, parameter, original samples), $\mathcal{L}_d$ denotes the dimension-specific loss function for measuring distillation quality.

\noindent \textbf{Selective Pixel Enhancement}
Effectively identifying and enhancing discriminative pixels is crucial for dataset quality, especially when addressing inter-class variability and non-discriminative regions.
Recent efforts have refined pixel-wise feature representation to emphasize informative pixels while suppressing redundant or misleading patterns.
Wang et al. \cite{wang2024emphasizing_arxiv2024} proposed the Emphasizing Discriminative Features (EDF) framework, which leverages Common Pattern Dropout (CPD) to filter out non-salient pixels and Discriminative Area Enhancement (DAE) to intensify attention on essential regions using Grad-CAM \cite{selvaraju2017grad_iccv2017} activation maps.
By adaptively selecting and enhancing discriminative regions, EDF ensures synthesized datasets prioritize high-salience regions, leading to improved generalization ability.
Extending this idea of emphasizing discriminative pixels, Tran et al. \cite{tran2025enhancingdatasetdistillationnoncritical_cvpr2025} proposed a region-based refinement strategy, which also uses CAM to precisely identify critical regions and preserve essential instance-specific details. 
In their method, non-critical regions are selectively refined to effectively incorporate broader class-general information into the distilled dataset.

\noindent \textbf{Selective Color Optimization}
Color redundancy often leads to inefficient dataset utilization, reducing the effectiveness of synthetic samples.
Yuan et al. \cite{yuancolor_nips2024} addressed this issue by introducing AutoPalette, a framework designed to selectively optimize color usage while preserving image features.
AutoPalette condenses images into reduced color palettes via a palette network, balancing pixel distributions with maximum color loss and palette balance loss.
A color-guided initialization strategy further ensures diverse, informative synthetic samples, improving overall distillation performance.
This method uniquely balances color diversity with feature preservation.
Compared to EDF \cite{wang2024emphasizing_arxiv2024} focusing on spatial region selection, AutoPalette targets redundancy within the color space, providing a complementary approach.
However, its effectiveness may be limited when
color is not a critical feature in datasets.

\noindent \textbf{Selective Parameter Optimization}
Conventional distillation methods often treat each network parameter equally, overlooking their varying contributions to the learning process.
Li et al. \cite{li2024importance_nn} highlighted this limitation and proposed the Adaptive Dataset Distillation (IADD) framework, which selectively assigns weights to parameters based on their contribution to learning.
This approach prioritizes critical parameters while maintaining challenging ones, resulting in improved performance and robust parameter matching across iterations.


\noindent\textbf{Sample Selection Strategies}
Dataset distillation methods often face significant challenges in effectively leveraging large-scale training data.
Sample selection methods tackle this problem through two main strategies: \textit{selection and pruning}, which reduce dataset size, and \textit{prioritization}, which dynamically emphasizes the most informative samples.

Selection and pruning methods aim to retain samples that best represent the underlying data distribution, enhancing the diversity and informativeness of the synthetic dataset.
Liu et al. \cite{liu2023dream_iccv2023} introduced DREAM, a clustering-based sample selection method designed to capture representative distributions, significantly enhancing training efficiency.
Building on this, DREAM+ \cite{liu2023dreamplus_arxiv2023} incorporates bidirectional matching to balance gradient and feature alignment, further improving stability and efficiency.
Moser et al. \cite{moser2024distill_arxiv2024} proposed the "Prune First, Distill After" framework, which combines dataset pruning with loss-value-based sampling.
This approach selects informative samples, reducing redundancy while maintaining performance.
Similarly, Bi-level Data Pruning (BiLP) \cite{xu2023distill_eccv2024} integrates preemptive pruning based on empirical loss and adaptive pruning leveraging causal effects, achieving up to significant data reduction with little degradation in performance.

Prioritization methods focus on dynamically ranking and weighting samples during training to emphasize their relative importance.
Wang et al. \cite{wang2024sdc_arxiv2024} introduced Sample Difficulty Correction (SDC), which prioritizes simpler samples based on gradient norms.
To adapt dataset difficulty dynamically, Lee et al. \cite{lee2024selmatch_icml2024} proposed SelMatch, which employs selection-based initialization and partial updates to fine-tune synthetic datasets.
By tailoring the training process to specific architectures and datasets, SelMatch improves generalization and convergence.
Extending the idea of curriculum-based sample prioritization, Chen et al. \cite{chen2025curriculumcoarsetofineselectionhighipc_cvpr2025} proposed a curriculum-based coarse-to-fine selection method to enhance high-IPC dataset distillation by adaptively selecting real training samples based on their difficulty. 
This method progressively excludes easy examples and carefully integrates challenging yet manageable instances, significantly improving the representativeness and generalization capability of distilled datasets.
Chen et al. \cite{chen2023dataset_arxiv2023} further extended prioritization by introducing an adversarial prediction matching framework.
This approach leverages teacher-student disagreements to identify informative samples, enhancing robustness across diverse architectures.
Tukan et al. \cite{tukan2023dataset_arxiv2023} proposed adaptive sampling and initialization strategies.
Focusing on data prioritization, Li et al. \cite{li2024prioritize_arxiv2024} proposed Prioritize Alignment in Dataset Distillation (PAD), improving distillation effectiveness through two key strategies: Data Filtering, which dynamically selects high-complexity samples based on EL2N scores \cite{paul2021deep_nips2021} to retain the most informative data, and Layer Confinement, which restricts the distillation process to deeper model layers, refining the prioritization of critical features while reducing noise from low-level representations.

\noindent \textbf{Stage Summary}
Sample selection methods enhance dataset distillation through two main approaches: reducing dataset size via selection and pruning, and optimizing training dynamics via prioritization.
Selection and pruning methods, such as DREAM+ and BiLP, reduce redundancy while preserving diversity and representational value.
In contrast, prioritization-based approaches, like SDC and SelMatch, dynamically adjust sample weights, improving generalization and training efficiency.
Future research could explore hybrid frameworks integrating various selection strategies for large-scale dataset distillation.

\subsubsection{Lossless Distillation}
Conventional dataset distillation methods often struggle with achieving lossless performance on complex datasets due to the challenges posed by various factors, such as inter-class variability and non-discriminative features. 
To address these issues, several recent approaches focus on achieving near-lossless or lossless performance by emphasizing critical patterns and refining trajectory alignment.

Guo et al. \cite{guo2024towards_iclr2024} laid the foundation for lossless distillation by introducing DATM, which dynamically aligns the difficulty of generated patterns with the synthetic dataset size. Specifically, early trajectories (easy patterns) are optimal for low-IPC settings, while late trajectories (hard patterns) benefit larger synthetic datasets. 
By controlling trajectory ranges and integrating soft label learning with sequential generation, DATM achieves near-lossless results across benchmarks such as CIFAR-10 and CIFAR-100.
Building on DATM, Zhou et al. \cite{zhou2024enhancing_arxiv2024} proposed M-DATM to address label inconsistencies and further optimize trajectory matching for more challenging datasets like Tiny ImageNet. 
Through deliberate refinements, M-DATM demonstrated superior performance, securing first place in the Fixed IPC Track at the ECCV-2024 Data Distillation Challenge.
Complementing trajectory-based approaches, Wang et al. \cite{wang2024emphasizing_arxiv2024} extended dataset distillation to high-variability datasets through feature importance optimization.  
By refining discriminative patterns, their method, EDF, demonstrated lossless performance on challenging subsets of ImageNet-1K, setting a new benchmark for dataset distillation.

\noindent \textbf{Stage Summary} These advancements collectively demonstrate the efficacy of adaptive selection strategies in lossless dataset distillation, where methods dynamically identify and prioritize the most informative and discriminative patterns while systematically minimizing the impact of redundant or less relevant information.
Moving forward, dynamic and context-aware adaptation techniques are expected to further enhance scalability and interpretability, enabling effective lossless distillation for large-scale datasets.

\subsubsection{Diversity of Distilled Data}
While importance-guided methods focus on selecting and prioritizing original samples, maintaining diversity in synthetic data remains a critical challenge in dataset distillation.
Recent approaches address this through curriculum learning, feature compensation, and semantic matching strategies.
Wang et al. \cite{ma2024curriculum_arxiv2024} introduced CUDD, which progressively increases synthetic data complexity through curriculum learning. 
By focusing on samples misclassified by student models, CUDD optimizes classification, regularization, and adversarial objectives to maintain diversity across training stages.
To address inter-class feature redundancy, Zhang et al. \cite{zhang2024breaking_arxiv2024} developed INFER with a Universal Feature Compensator (UFC). 
This "one instance for all classes" approach achieves superior performance on ImageNet through optimized feature diversity. 
Similarly, DELT \cite{shen2024delt_arxiv2024} enhances diversity through subtask partitioning with varying optimization schedules.
Du et al. \cite{Du2024DiversityDrivenSE_nips2024} observed limitations in instance-wise synthesis and proposed DWA (Directed Weight Adjustment), using variance-based regularization for batch-level diversity. 
DSDM \cite{li2024diversified_mm2024} further preserves semantic diversity by matching prototypes and covariance matrices of class distributions through pre-trained feature extractors.

\noindent \textbf{Stage Summary}  
These diversity-focused approaches complement original sample importance methods.  
While importance methods reduce redundancy at the original dataset level, diversity enhancement ensures richness at the distilled data level.  
Future research should explore hybrid frameworks that seamlessly integrate selection, prioritization, and diversity enhancement to create more generalizable, scalable dataset distillation pipelines.

\subsubsection{Augmentation Strategies}
Augmentation techniques expand feature and label spaces to enhance the learning dynamics of distillation, promoting greater diversity and representational richness.  
These methods ensure that synthetic data captures a broad range of features while remaining adaptable to different factors.  

\noindent \textbf{Model Augmentation}
Building on diversity-focused methods, model augmentation techniques expand feature spaces during dataset distillation, directly addressing the need for richer representations.
Zhang et al. \cite{zhang2023accelerating_cvpr2023} proposed integrating two model augmentation techniques within a gradient-matching framework: early-stage models and weight perturbation. 
Early-stage models provide a diverse feature space with larger gradient magnitudes, offering enhanced flexibility and richer guidance for the distillation process. 
Weight perturbation further expands the feature space by introducing a normalized random vector, sampled from a Gaussian distribution, to the model weights.
This augmentation fosters diversity by enabling the exploration of a broader range of feature representations, which, in turn, enhances the effectiveness of the distillation process.
Together, these techniques allow for both efficient distillation and the generation of more representative synthetic datasets.

\noindent \textbf{Label Augmentation}
Beyond feature spaces, label augmentation introduces diversity through enriched label representations.
Kang et al. \cite{Kang2024LabelAugmentedDD_arxiv2024} introduced Label-Augmented Dataset Distillation (LADD) to leverage label augmentation for improved diversity and efficiency. 
LADD operates in two stages: first, synthetic images are generated using existing distillation algorithms; second, an image sub-sampling algorithm generates multiple local views for each synthetic image. 
A pre-trained labeler then produces dense semantic labels for these local views. During deployment, LADD combines global view images with their original labels and local view images with the newly generated dense labels. 
This dual-label strategy enhances storage efficiency, reduces computational overhead, and provides diverse learning signals, leading to improved robustness and performance across a range of architectures.

\noindent \textbf{Stage Summary} The advancement of augmentation strategies in dataset distillation hinges on overcoming critical research challenges.  
A key open problem is the automatic discovery of optimal augmentation strategies, which could benefit from reinforcement learning approaches \cite{10172347_pami2023} to dynamically tailor augmentation techniques to specific tasks and model architectures.  Achieving this requires a deeper understanding of how different augmentation types interact with various model architectures, ensuring that strategies are not only effective but also generalizable and transferable across diverse domains and tasks.

\subsubsection{Extreme Compression Techniques}
Extreme compression techniques address storage constraints by drastically reducing dataset sizes while preserving training effectiveness.  
Shul et al. \cite{shul2024distilling_arxiv2024} introduced Poster Dataset Distillation (PoDD), which compresses entire datasets into a single poster image, achieving extreme compactness with less than one image per class.  
By leveraging semantic class ordering and efficient label management, PoDD balances high compression with robust training performance, making it well-suited for resource-constrained applications.  
Li et al. \cite{li2025contrastivelearningenhancedtrajectorymatching_arxiv2025} propose DATM-CLR to address the semantic degradation of synthetic data under extreme compression, e.g., IPC=1.
By integrating supervised contrastive learning into the image synthesis  stage, DATM-CLR enhances intra-class coherence and inter-class separability, achieving a significant accuracy gain over DATM on CIFAR-10 and consistent improvements across CIFAR-100 and Tiny-ImageNet.

\subsection{Distillation in Non-IID and Non-centralized Settings}
As machine learning systems increasingly deploy in dynamic environments, the traditional assumptions of independent and identically distributed (IID) data become increasingly inadequate. 
Modern applications that are ranging from edge computing and mobile networks to decentralized  systems, demand models that can adapt to heterogeneous, shifting data landscapes. 
Dataset distillation correspondingly evolve to address these complex challenges by handling non-IID data distributions, ensuring fairness and robustness in open-world scenarios, and accommodating decentralized learning architectures like federated learning.

\subsubsection{Addressing Non-IID Challenges}
Adapting dataset distillation methods to real-world applications requires addressing Non-IID challenges, including handling out-of-distribution (OOD) data \cite{yang2024generalized_IJCV2024}, mitigating biases \cite{jiang2022dataset_pami2022}, ensuring fairness \cite{huang2024federated_pami2024}, and supporting self-supervised \cite{zong2024self_pami2024,gui2024survey_tpami2024} and transfer learning \cite{5288526_kdd2010}. 
Recent advancements in these areas extend the applicability of dataset distillation beyond traditional IID settings, enabling robust performance in dynamic and decentralized environments.

\noindent \textbf{Out-Of-Distribution and Cross-Domain Generalization}
In open-world scenarios, models must reliably identify and handle data samples that deviate from their training distribution. Traditional dataset distillation methods overlook this, typically focusing only on IID data. 
To explicitly address the OOD detection problem within dataset distillation, Ma et al. introduced Trustworthy Dataset Distillation (TrustDD)~\cite{ma2025towards_pr2025}, integrating synthetic outlier generation via Pseudo-Outlier Exposure into the distillation process. 
TrustDD thereby eliminates dependence on curated OOD datasets, enabling robust OOD detection without sacrificing in-distribution accuracy or computational efficiency.  
Moving beyond single-domain scenarios, Choi et al.~\cite{choi2025damdomainawaremodulemultidomain_arxiv2025} addressed a related but distinct challenge: multi-domain dataset condensation. 
They proposed a Domain-Aware Module, employing frequency-based pseudo-domain labeling to automatically partition datasets into pseudo-domains, and learnable spatial domain masks to explicitly encode domain-specific features, significantly improving the synthesized dataset's generalization across diverse visual domains.

\noindent \textbf{Biased DD and Fairness DD}
Dataset biases, such as color or background amplification, can significantly undermine the representational quality of distilled datasets. 
Lu et al. \cite{lu2024exploring_cvpr2024} analyzed how biases propagate through the distillation process and proposed mathematical frameworks for addressing biased dataset distillation. 
Cui et al. \cite{cuimitigating_icml2024} introduced a reweighting scheme combining supervised contrastive learning and kernel density estimation to effectively mitigate these biases.
To ensure fairness in synthetic datasets, Zhou et al. proposed FairDD \cite{zhou2024fairdd_arxiv2024}, which aligns datasets with protected attribute groups through synchronized matching, mitigating majority group dominance. 
Similarly, Zhao et al. introduced Long-tailed Aware Dataset Distillation (LAD) \cite{zhao2024distilling_arxiv2024}, which addresses imbalances in long-tailed datasets by incorporating Weight Mismatch Avoidance and Adaptive Decoupled Matching. 
These methods demonstrate the potential for bias-aware dataset distillation to improve fairness and tail-class performance.

\noindent \textbf{Self-supervised and Transfer Learning}  
While most dataset distillation research has focused on supervised learning, self-supervised and transfer learning \cite{zong2024self_pami2024, 5288526_kdd2010,gui2024survey_tpami2024} remain underexplored.  
Lee et al. \cite{lee2023self_iclr2024} proposed KRR-ST, which employs kernel ridge regression and mean squared error objectives to reduce randomness in the distillation process. 
This method facilitates efficient pretraining with unlabeled data, significantly lowering computational costs.  
Expanding on this, Yu et al. \cite{yu2025self_iclr2025} introduced an enhanced self-supervised DD framework that builds upon KRR-ST through three key innovations: image and representation parameterization, predefined augmentation, and approximation networks. 
These refinements improve cross-architecture generalization, transfer learning performance, and storage efficiency.  
Furhter, Joshi et al. introduced MKDT \cite{joshi2024dataset_arxiv2024}, a two-stage method that stabilizes SSL trajectories by first training student models via knowledge distillation from SSL-trained teachers. 
In alignment with these stabilized trajectories, MKDT achieves substantial performance gains in SSL pretraining and downstream tasks.


\subsubsection{Addressing Non-centralized Challenges}
In the era of distributed machine learning, federated learning (FL) enables collaborative, privacy-preserving model training across decentralized environments. 
However, this paradigm introduces unique challenges, such as communication overhead, non-IID data, privacy concerns, and decentralized setups. 
Recent advancements in dataset distillation have addressed these challenges, categorized into communication efficiency, heterogeneity handling, privacy preservation, and decentralized/edge-focused scenarios.

\noindent \textbf{Communication Efficiency} 
Minimizing communication overhead is critical for scalable FL, especially in resource-constrained environments. 
Distilled One-Shot Federated Learning (DOSFL) \cite{zhou2020distilled_arxiv2020} introduced the idea of transmitting compact synthetic datasets in a single communication round, drastically reducing bandwidth usage. 
Building on this, FedSynth \cite{hu2022fedsynth_arxiv2022} replaced model updates with synthetic datasets, ensuring compatibility with standard FL frameworks. 
DENSE \cite{zhang2022dense_nips2022} and FedMK \cite{liu2022meta_iclr2023} avoided model aggregation by generating synthetic data that encapsulates local knowledge, while FedD3 \cite{song2023federated_ijcnn2023} optimized one-shot communication for edge scenarios. 
Further advancements, such as FedCache 2.0 \cite{pan2024fedcache_ap2024}, integrated distillation with knowledge caching to enhance efficiency in edge-device communication.

\noindent \textbf{Handling Heterogeneity}
Data and model heterogeneity pose fundamental challenges in federated learning, where clients may have vastly different data distributions, model architectures, and learning characteristics. 
FedDM \cite{xiong2023feddm_cvpr2023} and DYNAFED \cite{pi2023dynafed_cvpr2023} tackled data heterogeneity by generating compact pseudo datasets that align with local distributions. 
While FedDM iteratively refined synthetic datasets via distribution matching, DYNAFED leveraged early pseudo datasets with trajectory matching, prioritizing efficiency and privacy. 
FedAF (Aggregation-Free) \cite{wang2024aggregation_cvpr2024} mitigated client drift in non-IID settings by employing sliced Wasserstein regularization and collaborative condensation to harmonize client contributions. 
HFLDD \cite{shi2024dataset_arxiv2024} grouped clients into heterogeneous clusters, allowing cluster headers to aggregate distilled datasets and create approximately IID data, improving training efficiency in non-IID environments.

\noindent \textbf{Privacy Preservation}
Privacy concerns in FL have driven the integration of dataset distillation techniques that limit information sharing.
FedDGM \cite{jia2023unlocking_arxiv2023} utilizes pre-trained generators on the server to distill datasets into latent space, effectively preserving privacy by avoiding direct sharing of sensitive data. 
QUICKDROP \cite{dhasade2023quickdrop_arxiv2023} extended this concept by embedding dataset distillation into federated unlearning, reducing computational costs without sacrificing privacy or performance. 
Xu et al. \cite{xu2024flip_arxiv2024} applied the principle of least privilege (PoLP), where clients share only essential knowledge through local-global dataset distillation, further safeguarding sensitive information.
Moreover, FedWSIDD \cite{jin2025fedwsiddfederatedslideimage_arxiv2025} specifically addresses medical imaging scenarios, proposing a federated learning framework that exchanges synthetic slides instead of real patient data or model parameters, facilitating secure and efficient collaboration among diverse medical institutions.

\noindent \textbf{Decentralized and Edge-Focused Scenarios}
In fully decentralized setups and resource-constrained edge environments, new frameworks have been developed to tackle the challenges of operating \textit{without} a central server and the limitations of on-device training resources. 
DESA (Decentralized Federated Learning with Synthetic Anchors) \cite{huang2024overcoming_icml2024} introduced synthetic anchors—generated datasets approximating global data distributions—to align local features across clients, enabling knowledge sharing without a central server. 
For incremental learning in edge environments, Rub et al. \cite{rub2024continual_icps} proposed integrating dataset distillation with adaptive model sizing, meeting the challenges of TinyML and on-device training while maintaining performance.

\noindent \textbf{Others} Additional advancements include DKKT \cite{lee2024practical_arxiv2024}, which enriched distilled datasets using deep support vectors, achieving robust performance with just $1\%$ of the original dataset. 
Distributed Boosting (DB) \cite{chen2024distributed_cikm2024} enhanced the performance of prior methods through partitioning, soft-labeling, and integration strategies in distributed computing environments, setting new benchmarks on large-scale datasets such as ImageNet-1K. 
HFLDD \cite{shi2024dataset_arxiv2024} integrates dataset distillation into a federated framework by organizing clients into heterogeneous clusters. 
Within each cluster, clients transmit distilled datasets to cluster headers, which aggregate the distilled data to construct approximately IID datasets, significantly improving training efficiency and addressing non-IID data challenges.

\subsection{Robustness in Dataset Distillation}
As dataset distillation becomes increasingly applied in various scenarios, ensuring its robustness has become a critical focus. 
Synthetic datasets derived through distillation must be resilient to adversarial attacks \cite{wei2024physical_tpami2024}, backdoor threats \cite{li2022backdoor_tnnls2022}, and privacy breaches to ensure their reliability and effectiveness in real-world applications. 
However, those requirements introduce unique vulnerabilities; for example, small perturbations or manipulations can disproportionately affect their performance.
To address these challenges, recent advancements have introduced robust evaluation benchmarks, innovative defense strategies, and in-depth explorations of vulnerabilities in distillation pipelines.

\subsubsection{Adversarial Attack}
Adversarial robustness \cite{wei2024physical_tpami2024} in dataset distillation remains a relatively underexplored topic.
Recent studies have laid the groundwork for benchmarking and developing robust distillation techniques to enhance security.

\begin{figure}[t]
    \centering
\includegraphics[width=1.0\linewidth]{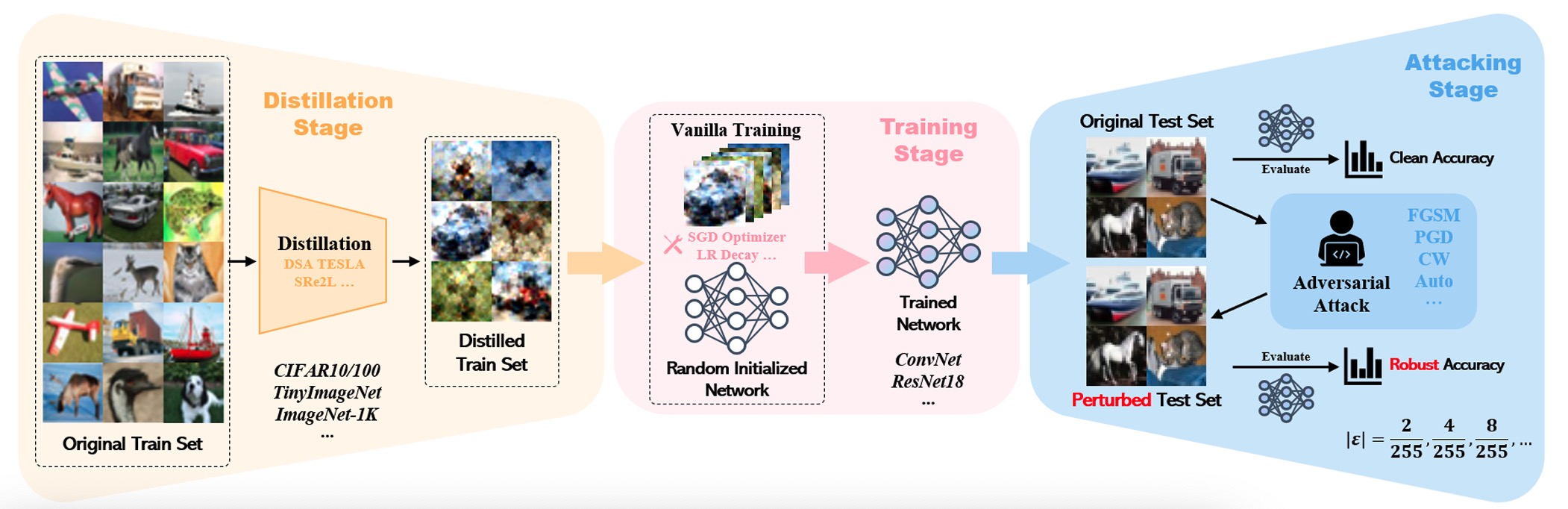}
    \caption{Overview of DD-RobustBench. Image from \cite{wu2024dd_arxiv2024}.}
    \label{fig:dd_robustbench}
\end{figure}

To assess the vulnerability of dataset distillation, Wu et al. \cite{wu2024dd_arxiv2024} introduced DD-RobustBench, a comprehensive benchmark for evaluating the robustness of dataset distillation methods, including techniques like TESLA, DREAM, SRe2L, and D4M, across datasets such as CIFAR-10 and ImageNet-1K. 
Their findings revealed that models trained on distilled datasets could exhibit superior robustness compared to those trained on original data under low IPC (images per class) settings, though an inverse relationship between robustness and IPC challenged the assumption that more distilled images always improve security. 
Following this, Zhou et al. \cite{zhou2024beard_arxiv2024} proposed BEARD, a unified evaluation framework introducing metrics like Robustness Ratio (RR), Attack Efficiency Ratio (AE), and Comprehensive Robustness-Efficiency Index (CREI) to assess robustness and efficiency holistically. 
Together, these benchmarks provide critical insights into balancing robustness and efficiency in adversarially robust distillation.
Together, these benchmarks establish a foundation for evaluating and improving the security of distillation techniques.

Beyond benchmarking, recent methods have focused on actively enhancing the adversarial robustness of distilled datasets. 
Xue et al. \cite{xue2024towards_arxiv2024} proposed Geometric regUlarization for Adversarial Robust Dataset (GUARD), a novel method incorporating curvature regularization to improve resilience against adversarial attacks. 
By minimizing the curvature of the loss function, GUARD reduces the upper bound of adversarial loss without significant computational overhead. 
This method achieves enhanced adversarial robustness with minimal computational overhead, paving the way for more secure and reliable distillation processes.

\subsubsection{Backdoor Attack}
Dataset distillation has been revealed to be vulnerable to backdoor attacks \cite{li2022backdoor_tnnls2022}, where attackers can embed hidden triggers that cause targeted misclassifications even after the dataset distillation process \cite{liu2023backdoor_ndss}.
To handle this challenge, Chung et al. \cite{chung2023rethinking_arxiv2023} analyzed backdoor attacks using a kernel method framework, explaining the resilience of certain backdoors in distilled datasets. 
They proposed two novel trigger generation strategies: simple-trigger, which exploits the observation that larger triggers reduce the generalization gap, and relax-trigger, which minimizes both conflict and projection losses while maintaining small generalization gaps. 
Empirical evaluations demonstrated that both methods evade existing backdoor detection and defense techniques effectively, with relax-trigger exhibiting superior robustness across eight tested defenses.

Extending backdoor attack risks beyond Euclidean data, Wu et al. \cite{wu2024backdoor_arxiv2024} introduced the first backdoor attack framework for graph condensation, named BGC. 
BGC injects malicious triggers into graph structures and optimizes them iteratively throughout the condensation process. 
To maximize attack effectiveness under constrained resources, BGC employs a representative node selection mechanism to identify and poison key nodes strategically. 
Experiments confirmed BGC’s high attack success rates and robust performance, even against multiple defense strategies, highlighting the critical need for developing secure graph condensation.

\subsubsection{Beyond Adversarial and Backdoor Attacks}
While adversarial and backdoor attacks represent critical security concerns in dataset distillation, recent studies have uncovered additional vulnerabilities that demand attention. 
These include privacy-related threats such as membership inference attacks, which jeopardize the confidentiality of the training process and expose sensitive data.
Chen et al. \cite{chen2023comprehensive_arxiv2023} conducted the first comprehensive study on the security properties of dataset distillation methods, specifically analyzing their vulnerability to membership inference attacks \cite{hu2022membership}. 
These attacks expose significant privacy concerns, as adversaries can infer whether a particular sample was used in the training process.

\subsection{Model-agnostic Solutions}
A major challenge in dataset distillation is the limited cross-architecture generalization capability of synthetic datasets. 
This issue arises when the distillation process becomes overly specialized to the architecture used during training, limiting the versatility of distilled datasets. 
Recent efforts have tried addressing this limitation by developing model-agnostic solutions that enhance the adaptability and transferability of synthetic datasets across diverse architectures.

\noindent \textbf{Gradient Balancing and Semantic Alignment}
Moon et al. \cite{moon2024towards_eccv2024} introduced Heterogeneous Model Dataset Condensation (HMDC), a pioneering approach to creating universally applicable condensed datasets. 
HMDC tackles the gradient imbalance problem, where contributions from different architectures vary due to gradient magnitude discrepancies, by employing a Gradient Balance Module (GBM). This module normalizes gradients to ensure equal contributions from heterogeneous models. 
To address misalignment, HMDC incorporates Mutual Distillation (MD) with Spatial-Semantic Decomposition (SSD), aligning both semantic and spatial features across models. 
These innovations ensure consistent feature representation and robust generalization, making HMDC a benchmark for cross-architecture adaptability.

\noindent \textbf{Mitigating Inductive Bias}
To combat architecture-specific inductive biases, Zhao et al. proposed two complementary methods: ELF \cite{zhao2023boosting_arxiv2023} and MetaDD \cite{zhao2024metadd_arxiv2024}. ELF leverages external supervision by using bias-free intermediate features extracted from original datasets to guide the training of evaluation models. 
In contrast, MetaDD takes an internal approach, disentangling features into architecture-invariant meta features and architecture-variant heterogeneous features. 
This ensures that generalizable features dominate the distillation process. While ELF emphasizes external supervisory signals to counteract overfitting, MetaDD focuses on directly optimizing the feature space. 
Together, these approaches significantly enhance the cross-architecture generalization of synthetic datasets.

\noindent \textbf{Architectural Diversity and Ensemble Techniques}
Recent advances have focused on leveraging multiple architectures to enhance robustness. 
Zhou et al. \cite{zhou2024improve_arxiv2024} proposed a model pool framework to balance architectural diversity and training stability. 
By combining multiple similar architectures and selecting a primary model most of the time, this method ensures robust performance across various architectures. 
Knowledge distillation further aligns student and teacher models, improving generalization.
Similarly, Zhong et al. \cite{zhong2023towards_arxiv2023} addressed architectural overfitting with a set of complementary techniques. 
These include a DropPath variant for implicit subnetwork ensembling, reverse knowledge distillation where smaller models teach larger ones, and optimizations like periodic learning rate schedules and advanced data augmentations. 

\noindent \textbf{Stage Summary} While these works have made significant progress in addressing cross-architecture generalization,  challenges persist in the field. 
The scalability to larger and more diverse architectures, such as Transformers \cite{han2022survey_tpami2022}, remains a crucial concern. 
Future research in this area might explore adaptive architecture-aware distillation strategies that dynamically adjust to different model architectures.

\section{Emerging Applications and Domains}
Dataset distillation has extended beyond image classification to complex data types such as temporal sequences, multi-modal data, and domain-specific challenges like medical imaging. 
These advancements underscore its adaptability across fields, including video understanding, natural language processing, and multi-modal learning. 
The following subsections summarize key applications, demonstrating how distillation techniques tackle domain-specific challenges while leveraging core methodologies.

\subsection{Temporal Domain}
The temporal domain introduces unique challenges for dataset distillation due to the sequential and time-dependent nature of video and audio data.
Recent advancements have focused on designing frameworks that effectively capture temporal dynamics, ensuring robust representation and efficient compression across these modalities.

\noindent \textbf{Video Domain}
Wang et al. \cite{wang2024dancing_cvpr2024} conducted the first systematic study on video dataset distillation, classifying temporal compression methods into four dimensions: synthetic frames, real frames, segments, and interpolation algorithms. 
They proposed a two-stage framework that disentangles static and dynamic information, integrating single-frame static memory with dynamic memory blocks to synthesize realistic video data. 
Building on this, Chen et al. \cite{Chen2024ALS_arxiv2024} introduced an efficient temporal condensation strategy incorporating slide-window sampling and adaptive trajectory matching, significantly improving performance on action recognition benchmarks while  reducing training costs.
Most recently, Li et al. \cite{li2025latent_cvprw2025} proposed the first latent-space video dataset distillation framework, combining VAE-based video encoding with high-order singular value decomposition (HOSVD) for compression. 
Leveraging latent representations introduced significant theoretical advancements, though practical implications for video quality merit further exploration.
Furthermore, Li et al. \cite{li2025videodatasetcondensationdiffusion_arxiv2025} proposed a novel video dataset distillation framework leveraging a pretrained video diffusion model, Latte \cite{ma2024latte_arxiv2024}, to efficiently generate high-quality synthetic videos. 
Subsequently, two innovative video selection strategies were introduced: (1) VST-UNet, a 4D spatio-temporal U-Net architecture optimized via cross-entropy, diversity, and representativeness losses, and (2) TAC-DT, a training-free, temporal-aware clustering method utilizing VideoMAE embeddings and hierarchical BIRCH clustering. 
Employing diffusion models and advanced selection strategies demonstrated substantial benefits, marking a promising future direction in the field.

\noindent \textbf{Audio Domain} Jiang et al. \cite{jiang2024ddfad} introduced DDFAD for audio data, leveraging Fused Differential MFCC (FD-MFCC) to combine traditional MFCC features with first- and second-order derivatives, improving feature richness. 
Using the Matching Training Trajectory  technique, DDFAD achieves performance comparable to full datasets while reducing computational and storage demands.

\noindent \textbf{Sequential Data}
Zhang et al. \cite{zhang2025td3_www2025} introduce Tucker Decomposition based Dataset Distillation (TD3), a novel Tucker Decomposition-based framework for dataset distillation in sequential recommendation, demonstrating an innovative approach to generating compact yet expressive synthetic sequence summaries.

\subsection{Multi-modal Dataset Distillation}
Recent advancements have extended dataset distillation to multi-modal tasks, addressing the unique challenges of integrating vision, language, and audio modalities. Vision-language learning \cite{wu2024visionlanguage,xu2024low_icml2024} and audio-visual learning \cite{kushwaha2024audiovisual_tmlr2024} have emerged as prominent areas of exploration.
Wu et al. \cite{wu2024visionlanguage} introduced a bi-trajectory matching framework for vision-language distillation, aligning image-text correspondences using contrastive loss and trajectory matching. 
By incorporating Low-Rank Adaptation (LoRA) \cite{hu2021lora_arxiv2021}, this method enhances efficiency while maintaining robust cross-modal representations.
Complementing this, Xu et al. \cite{xu2024low_icml2024} proposed Low-Rank Similarity Mining (LoRS), which distills similarity matrices alongside image-text pairs. 
Using low-rank factorization, LoRS efficiently approximates similarity matrices while refining anchors for contrastive learning, improving representation and compression.

Expanding on these developments, Kushwaha et al. \cite{kushwaha2024audiovisual_tmlr2024} introduced audio-visual dataset distillation, a technique that compresses large audio-visual datasets into compact synthetic datasets while preserving cross-modal relationships. 
Their framework builds on vanilla distribution matching by incorporating two novel losses: implicit cross-matching and cross-modal gap matching, ensuring better alignment between synthetic and real data.

Most recently, Zhang et al.~\cite{zhang2025modalitycollapserepresentationsblending_arxiv2025} provide a systematic diagnosis of why prior multimodal dataset distillation methods struggle. 
They identify a fundamental conflict between over-compression (inherent to dataset distillation) and contrastive supervision (essential for cross-modal alignment), which leads to {modality collapse}—a phenomenon where intra-modal representations become overly concentrated while inter-modal divergence increases. 
To address this, they propose {RepBlend}, a lightweight yet effective framework that mitigates modality collapse through {representation blending} to enhance intra-modal diversity and {symmetric trajectory matching} to balance supervision across modalities. 
Beyond performance gains, RepBlend also highlights future research directions such as fine-grained cross-modal alignment.

\subsection{Medical Domain}  In medical imaging, Li et al. \cite{li2024dataset_arxiv2024} examined dataset distillation’s feasibility across nine diverse medical datasets, demonstrating its ability to preserve diagnostic details while reducing dataset sizes. 
They observed that larger inter-class variations yield better results and proposed random selection as a heuristic predictor of distillation efficacy. 
Yu et al. \cite{yu2024progressive_arxiv2024} addressed training instability caused by SGD oscillations with a progressive trajectory matching strategy. 
By gradually increasing trajectory step size and introducing a dynamic overlap mitigation module with Maximum Mean Discrepancy \cite{smola2006maximum}, Yu et al. achieved stable training and notable performance improvements, especially under low IPC conditions. 
Complementing these efforts, Li et al. \cite{li2024image_miccai2024} proposed InfoDist, a privacy-preserving framework using class-conditional latent diffusion models to generate synthetic histopathology datasets. 
By selecting informative images based on modular centrality and enhancing performance through contrastive learning, InfoDist safeguards data privacy while ensuring competitive accuracy.

\subsection{Other Applications} 
\noindent \textbf{Scientific Discovery} Dataset distillation has demonstrated its versatility in tackling scientific challenges, such as galaxy morphology analysis \cite{guan2023discovering_arxiv2023}.
Guan et al. \cite{guan2023discovering_arxiv2023} applied dataset distillation to galaxy morphology analysis, proposing Self-Adaptive Trajectory Matching (STM) to improve upon MTT. 
STM uses statistical hypothesis testing to adaptively monitor validation loss, ensuring efficient stopping criteria and reduced hyperparameter tuning.
Their work includes a curated high-confidence version of the Galaxy Zoo 2 dataset \cite{willett2013galaxy}, successfully distilling morphological features and achieving superior performance.

\noindent \textbf{Object Detection and Image Restoration} 
For object detection, Qi et al.~\cite{qifetch_nips2024} introduced DCOD, the first dataset condensation framework specifically tailored to this task, demonstrating competitive performance on standard benchmarks such as Pascal VOC~\cite{hoiem2009pascal} and MS COCO~\cite{lin2014microsoft_eccv2014}. 
Extending dataset distillation beyond detection, recent approaches have also addressed challenging image restoration tasks. 
For instance, Dietz et al.~\cite{Dietz2025ASI_arxiv2025} proposed a distillation framework focusing on adaptive super-resolution tasks, while Peng et al.~\cite{peng2025instancedatacondensationimage_arxiv2025} developed an instance-aware distillation method aimed explicitly at improving super-resolution performance. 
In parallel, Zheng et al.~\cite{zheng2025distributionawaredatasetdistillationefficient_arxiv2025} introduced TripleD, a distribution-aware dataset distillation framework, explicitly designed to optimize image restoration tasks by better preserving underlying data distributions.

\noindent \textbf{Enhancing Data Quality} Dataset distillation has  shown promise in noise-related challenges and improving dataset quality.
Cheng et al. \cite{cheng2024dataset_arxiv2024} proposed leveraging dataset distillation as a novel denoising tool for learning with noisy labels, addressing challenges like feedback loops in traditional noise evaluation strategies. 
Through extensive experiments with methods like DATM, DANCE, and RCIG, they demonstrate the effectiveness of dataset distillation in removing random and natural noise, though it struggles with structured asymmetric noise. 
This approach also improves training efficiency and privacy by avoiding direct use of noisy data and enabling offline processing.
Building upon the insights into noisy label handling through distillation, Wu et al. \cite{wu2025trustawarediversiondataeffectivedistillation_arxiv2025} proposed Trust-Aware Diversion (TAD). 
Rather than just using distillation as a denoising tool \cite{cheng2024dataset_arxiv2024}, TAD introduces a dual-loop optimization framework that actively manages noisy labels during the distillation process itself.
In TAD, an outer loop  separates data into trusted and untrusted spaces to guide distillation, while an inner loop recalibrates untrusted samples to maximize their utility. 
UniDetox \cite{lu2025unidetox_iclr2025} utilized dataset distillation approach to universally mitigate toxic content in large language models by systematically refining training datasets, thereby reducing harmful biases while preserving model performance across different domains.

\section{Performance Comparison}
Dataset distillation methods have advanced significantly, as demonstrated in Table \ref{tab:survey_comparison}, which compares state-of-the-art techniques across CIFAR-10, CIFAR-100, Tiny ImageNet, and ImageNet-1K under different IPC settings (IPC=1,10,50,100,200), as well as Table \ref{tab:single-dataset-comparison}, which compares state-of-the-art techniques across ImageNet-21K under different IPC settings.
These evaluations provide insights into efficiency and scalability, offering a comprehensive view of dataset distillation capabilities.
Over the past two years, the field has seen rapid progress, particularly in low-IPC scenarios, where newer approaches substantially improve upon early methods. For instance, on CIFAR-10 with IPC=1, early techniques such as DC \cite{zhao2020dataset_iclr2021} achieved only 28.3\% accuracy, whereas recent advances like AutoPalette \cite{yuancolor_nips2024} have pushed this performance to 58.6\%, demonstrating a significant 30.3\% absolute improvement. Similar trends are observed across larger-scale datasets. 
On CIFAR-100, the performance has improved from 12.8\% (DC) to 38.0\% (AutoPalette) with IPC=1, while on ImageNet-1K, recent methods like D4M \cite{su2024d_cvpr2024} achieve 66.5\% accuracy with IPC=100, showcasing the scalability of modern distillation approaches.
Furthermore, the emergence of diverse methodologies, including SRe2L, Selective-based, and Diversity-driven approaches, has contributed to these improvements. 
As shown in Table \ref{tab:single-dataset-comparison}, even on the challenging ImageNet-21K dataset, modern methods like CUDD \cite{ma2024curriculum_arxiv2024} achieve promising results, reaching 34.9\% accuracy with IPC=20. 
These advancements demonstrate the field's progress in handling increasingly complex and large-scale datasets while maintaining efficiency through low IPC settings.

\subsection{Impact of IPC Settings and Practical Considerations}
The relationship between IPC settings and model performance exhibits distinct patterns across different scales of datasets and methodologies. Analysis of recent results reveals three characteristic regions of IPC impact:

(1) Low IPC regime (IPC $\leq$ 10) demonstrates the most dramatic improvements in performance. 
On CIFAR-10, D3M \cite{abbasi2024one_arxiv2024} shows a substantial improvement from 35.9\% (IPC=1) to 58.6\% (IPC=10), while DANCE \cite{Zhang2024_ijcai2024} improves from 47.1\% to 70.8\%. 
Similar patterns are observed on CIFAR-100, where methods like EDC \cite{shao2024elucidating_nips2024} achieve significant gains from IPC=1 to IPC=10.
(2) Mid-range IPC values (10-50) show continued but moderated improvements. For instance, INFER \cite{zhang2024breaking_arxiv2024} on CIFAR-100 improves from 50.2\% (IPC=10) to 65.1\% (IPC=50). 
This range often represents the optimal trade-off between compression and performance for practical applications.
(3) High IPC settings (50-100) exhibit diminishing returns across datasets. On CIFAR-100, PAD \cite{li2024prioritize_arxiv2024} shows only marginal gains from IPC=50 (55.9\%) to IPC=100 (58.5\%). 
This pattern is consistently observed in larger datasets like ImageNet-1K, as demonstrated by methods such as CDA \cite{yin2023dataset_arxiv2023} and G-VBSM \cite{shao2024generalized_cvpr2024}.

The systematic evaluation of high IPC settings (IPC$>$50) on large-scale datasets such as ImageNet-1k and ImageNet-21K was largely unfeasible two years ago due to computational and methodological constraints. 
Recent advances have overcome these limitations, enabling comprehensive analysis of IPC scaling effects across diverse datasets and methods. 
For example, methods like D4M \cite{su2024d_cvpr2024} and DELT \cite{shen2024delt_arxiv2024} now demonstrate strong performance even at high IPC values on ImageNet-1K, with D4M achieving 66.5\% accuracy at IPC=100, 68.1\% accuracy at IPC=200. 
While higher IPC values can achieve better performance, the increased storage and computational requirements may outweigh the marginal gains in many practical scenarios. 
For efficient pretraining and real-world applications, IPC values between 10 and 50 typically offer the most practical balance of performance and resource efficiency.

\subsection{Performance Analysis Across Dataset Scales and Complexities}
Performance varies significantly across datasets, revealing clear trends as dataset scale and complexity increase.
Recent methods have achieved remarkable progress on CIFAR-10, with several approaches exceeding 50\% accuracy at IPC=1, a substantial improvement over earlier methods that reached around 30\%.
However, this success diminishes as complexity grows, such as in the transition from CIFAR-10 to CIFAR-100, where the increased class number and greater inter-class variability pose significant challenges

Quantitative analysis reveals consistent performance degradation across dataset scales. 
For instance, DATM \cite{guo2024towards_iclr2024} achieves 46.9\% accuracy on CIFAR-10 with IPC=1 but drops to 27.9\% on CIFAR-100. 
Similarly, RDED \cite{sun2024diversity_cvpr2024} shows a dramatic decrease from 22.9\% on CIFAR-10 to 11.0\% on CIFAR-100 at IPC=1. 
This pattern becomes more pronounced with larger-scale datasets.
DATM's performance drops from 66.8\% on CIFAR-10 to 31.1\% on Tiny ImageNet at IPC=10, highlighting the challenges of preserving semantic information under limited IPC settings as the complexity increases.

The transition to high-resolution, large-scale datasets like ImageNet-1K and ImageNet-21K introduces challenges beyond increased class counts. 
These datasets feature more complex visual characteristics, including varied perspectives and extensive intra-class variations. 
Recent methods address these challenges through innovative approaches. 
For example, INFER \cite{zhang2024breaking_arxiv2024} and DWA \cite{Du2024DiversityDrivenSE_nips2024} leverage diversity-driven strategies and advanced regularization techniques to enhance representational quality. 
On ImageNet-1K, these methods demonstrate promising scalability, with DWA achieving 55.2\% accuracy at IPC=50 and INFER showing robust performance across different IPCs.

The scaling to ImageNet-21K presents an even greater challenge with its massive $21,000$ classes. 
Nevertheless, recent methods have shown encouraging results. CUDD \cite{ma2024curriculum_arxiv2024} achieves 34.9\% accuracy at IPC=20, while CDA \cite{yin2023dataset_arxiv2023} reaches 26.4\% at the same IPC setting. 
Even with limited IPC (IPC=10), methods like EDC \cite{shao2024elucidating_nips2024} and RDED \cite{sun2024diversity_cvpr2024} maintain reasonable performance at 26.8\% and 25.6\% respectively. 
These results on ImageNet-21K, though lower than those on smaller datasets, represent significant progress in scaling dataset distillation to extremely large-scale scenarios.

These advancements represent significant progress in scaling dataset distillation to complex, real-world scenarios. 
However, the persistent performance gap between smaller and larger datasets indicates that maintaining distillation quality across varying dataset scales remains a key challenge in the field. 
The success of diversity-focused and regularization-enhanced approaches suggests promising directions for future research in handling large-scale datasets.

\begin{table*}[ht]
    \centering
    \caption{Performance comparison of dataset distillation methods across four datasets (CIFAR-10/100, Tiny ImageNet, and ImageNet-1K) under different IPCs. R18 denotes ResNet18 architecture. Methods without explicit R18 notation use ConvNet as the default architecture.}\label{tab:performance}
    \rowcolors{2}{rowcolor1}{rowcolor2}
    \tiny
        \begin{tabular}{|l|c|c|cccc|cccc|cccc|ccccc|}
        \hline
                \rowcolor{white}\global\rownum=\numexpr\rownum-1\relax
     \multirow{2}{*}{\textbf{Methods}} & \multirow{2}{*}{\textbf{Schemes}} & 
     \multirow{2}{*}{\textbf{Venue}} &
     \multicolumn{4}{c|}{\textbf{CIFAR-10}} & \multicolumn{4}{c|}{\textbf{CIFAR-100}} & \multicolumn{4}{c|}{\textbf{Tiny ImageNet}} & \multicolumn{5}{c|}{\textbf{ImageNet-1K}}   \\
        & & &1& 10 & 50 & 100 & 1& 10 & 50 & 100 & 1& 10 & 50& 100 & 1& 10 & 50 & 100 &200  \\
        \hline
        DD \cite{wang2018dataset_arxiv2018} & META & arXiv/2018 & 42.8 & - & - & - & - & - & - & - & - & - & - & -& - & - & -& -& -\\ 
        RaT-BPTT \cite{feng2023embarrassingly_iclr2023} & META& ICLR/2023 & 53.2 & 69.4 & 75.3 & - & 35.3 & 47.5 & 50.6 & - & 20.1 & 24.4 & - & -& - & - & -& -& -\\
        Teddy \cite{yu2025teddy_eccv2024} & META& ECCV/2024 & 30.1 & 53.0 & 66.1 & - & 13.5& 33.4 & 49.4 & - & - & - & 45.2 & 52.0& - & 34.1 & 52.5& 56.5& -\\ \hline
        DC \cite{zhao2020dataset_iclr2021} & GM& ICLR/2021 & 28.3 & 44.9 & 53.9 & - & 12.8 & 25.2 & - & - & - & - & 11.2 & -& - & - & -& -& -\\
        DSA \cite{zhao2021dataset_icmlr2021} & GM& ICML/2021 & 28.8 & 52.1 & 60.6 & - & 13.9& 32.3 & - & - & - & - & 25.3 & -& - & - & -& -& -\\ \hline
        MTT \cite{cazenavette2022dataset_cvpr2022} & TM& CVPR/2022 & 46.3 & 65.3 & 71.6 & - & 24.3& 40.1 & 47.7 & - & 8.8 & 23.2 & 28.0 & 33.7& - & - & -& -& -\\
        FTD \cite{du2023minimizing_cvpr2023} & TM& CVPR/2023 & 46.8 & 66.6 & 73.8 & - & 25.2& 43.4 & 50.7 & - & 10.4 & 24.5 & - & -& - & - & -& -& -\\
        TESLA \cite{cui2023scaling_icml2023} & TM& ICML/2023 & 48.5 & 66.4 & 72.6 & - & 24.8& 41.7 & 47.9 & - & - & - & - & -& 7.7 & 17.8 & 27.9& -& -\\
        AST \cite{shen2023ast_arxiv2023} & TM& arXiv/2023 & 48.8 & 67.1 & 74.6 & - & 26.6& 44.4 & 51.7 & - & 13.7 & 25.7 & - & -& - & - & -& -& -\\
        Li et al. \cite{li2024dataset_ieice2024} & TM& IEICE/2024 & 46.4 & 65.5 & 71.9 & - & 24.6& 43.1 & 48.4 & - & - & - & - & -& - & - & -& -& -\\
        ATT \cite{liu2024dataset_eccv2024} & TM& ECCV/2024 & 48.3 & 67.7 & 74.5 & - & 26.1 & 44.2  & 51.2 & - & 11.0 & 25.8 & - & -& - & - & -& -& -\\
        MCT \cite{zhong2024towards_arxiv2024} & TM& CVPR/2025 & 48.5 & 66.0 & 72.3 & - & 24.5& 42.5 & 46.8 & - & 9.6 & 22.6 & 27.6 & -& - & - & -& -& -\\
        MTT-LSS \cite{kong2025efficient_icassp2025} & TM& ICASSP/2025 & 65.3 & 71.3 & 74.8 & - & 33.9& 46.3 & 50.0 & - & - & - & - & -& - & - & -& -& -\\
        \hline
        CAFE \cite{wang2022cafe_cvpr2022}& DM& CVPR/2022 & 30.3 & 46.3 & 55.5 & - & 12.9  &  27.8 & 37.9 & - & - & - & - & -& - & - & -& -& -\\
        DM \cite{zhao2023dataset_wacv2023}  & DM& WACV/2023 & 26.0 & 48.9 & 63.0 & - & 11.4  &  29.7 & 43.6 & - & 3.9 & 12.9 & 24.1 & -& 1.3 & 5.7 & 11.4& -& -\\
        DataDAM \cite{sajedi2023datadam_iccv2023} & DM& ICCV/2023 & 32.0 & 54.2& 67.0 & - & 14.5  &  34.8 & 49.4 & - & 8.3 & 18.7 & 28.7 & -& 2.0 & 6.3 & 15.5& -& -\\ 
        IDM \cite{zhao2023improved_cvpr2023} & DM& CVPR/2023 & 45.6 & 58.6 & 67.5 & - & 20.1  &  45.1 & 50.0 & - & 10.1 & 21.9 & 27.7 & -& & -- & - & -& -\\
        WMDD \cite{liu2023dataset_arxiv2023}  & DM& arXiv/2023 & - & -& - & - & -  &  - & - & - & 7.6& 41.8 & 59.4 & 61.0 & 3.2 & 38.2 & 57.6& 60.7& -\\
        Rahimi et al. \cite{Malakshan2024distilling_arxiv2024} & DM& arXiv/2024 & 27.9 & 53.0& 65.6 & - & 13.5  &  33.9 & 45.3 & - &4.9 & 17.2 & 27.4 &  - &2.1 & 7.5 & 15.6 & -& -\\
        DANCE \cite{Zhang2024_ijcai2024}  & DM& IJCAI/2024 & 47.1 & 70.8& 76.1 & - & 27.9  &  49.8 & 52.8 & - &11.6 & 26.4 & 28.9 &  - &- & - & -& - & -\\
        M3D \cite{zhang2024m3d_aaai2024} & DM& AAAI/2024 & 45.3 & 63.5& 69.9 & - & 26.2  &  42.4 & 50.9 & - & - & - & -& - & -& - & - & -& -\\ 
        LQM \cite{wei2024dataset_cvprw2024} & DM& CVPRW/2024 & 45.9 & 60.9& 70.2 & - & 27.2  &  47.7 & 52.4 & - & 10.4 & 20.8 & 24.3 & -& - & - & -& -& -\\
        Deng et al. \cite{deng2024exploiting_cvpr2024} & DM& CVPR/2024 & 47.1 & 59.9& 69.0 & - & 24.6  &  45.7 & 51.3 & - &10.0 & 23.3 & 27.5 &  - &- & - & - & -& -\\
        NCFM \cite{wang2025dataset_cvpr2025} & DM& CVPR/2025 & 49.5 & 71.8& 77.4 & - & 34.4  &  48.7 & 54.7 & - &18.2 & 26.8 & 29.6 &  - &- & - & - & -& -\\
        CaO2~\cite{wang2025_iccv2025} & DM& ICCV/2025 & - &- & - & - & - &- & - & - &- &- & -&  - &7.1 & 46.1 & 60.0 & -& -\\
        Zou et al.~\cite{zou2025_iccv2025} & DM& ICCV/2025 & - &39.0 & 63.2 & - & - &50.6 & 66.1 & - &- &- & -&  - &- & 46.7 & 60.5 & -& -\\
        \hline
        FreD \cite{shin2024frequency_nips2023} & Latent& NeurIPS/2023 & 60.6 & 70.3& 75.8 & - & 34.6  &  42.7 & 47.8 & - &- & - & - &  - &- & - & - & -& -\\ 
        NSD \cite{yang2024neural_eccv2024}  & Freq& ECCV/2024 & 68.5 & 73.4 & 75.2 & - & 36.5  &  46.1 & - & - & 21.3 & - & - & -& - & - & -& -& -\\
        CFM \cite{bo2025understandingdatasetdistillationspectral_arxiv2025}  & Freq/R18& arXiv/2025 & - & 57.0 & 82.3 & - & -  &  64.6 &71.4 & - & - & -  & 54.6& 57.6& - & 38.3 & 57.3& 60.1& 63.6\\
\hline
MIM4DD  \cite{shang2024mim4dd_nips2023} &PP& NeurIPS/2023 & 51.9 & 70.8 & 74.7& - & 31.1 & 47.4  &  - & - & -&-  & - &  - &- & - & -& - & -\\
SeqMatch \cite{du2024sequential_nips2024} &PP& NeurIPS/2024 & - & 68.3 & 75.3& - & - & 45.1  &  51.9 & - & -&23.8  & - &  - &- & -& - & - & -\\
FYI \cite{son2024fyi_eccv2024} &PP& ECCV/2024 & 52.5 & 68.2 & 75.2& - & 28.9& 45.8 & 50.8  &  -  & 11.6&26.8  & 30.1 &  - &- & - & - & -& -\\
CMI \cite{zhong2024going_arxiv2024} &PP& ICLR/2025 & - & 70.0 & 76.6& - & -& 46.6 & 53.8  &  -  & 10.4&25.7  & 30.1 &  - &- & 24.2 & 49.1 & 54.6& -\\
H-GLaD (MTT) \cite{zhong2024hierarchical_cvpr2025} &PP& CVPR/2025 & 37.2 & - & -& - & -& - & -& - & -&- & -& - & - & - & -& - & -\\
OPTICAL \cite{cui2025optical_cvpr2025} &PP& CVPR/2025 & 52.0 & 71.1 & 76.5& - & 29.7& 50.7 & 53.5& - & 12.3&26.9 & 29.6& - & - & - & -& - & -\\
\hline
DiM \cite{wang2023dim_arxiv2023} &Gen& arXiv/2023 & 51.3 & 66.2 & 72.6& - & - & - & -& -& - & - & -& - &- & - & - & -& -\\
Zhang et al. \cite{zhang2023dataset_arxiv2023}&Gen& arXiv/2023 & 48.2 & 66.2 & 73.8& - & 26.1 & 41.9 & 48.5& -& - & - & -& - &7.9 & 17.6 & 27.2 & -& -\\
D3M \cite{abbasi2024one_arxiv2024}&Gen& arXiv/2024 & 35.9 & 58.6 & 70.5& - & 30.8 & 49.1 & 54.5& -& 11.4 & 38.8 & 51.4 & -& 5.0 &23.6 & 32.2 & -& -\\
Li et al. \cite{li2024generative_cvpr2024} &Gen& CVPRW/2024 & 52.3 & 66.7 & 73.1& - & - & - & -& -& - & - & -& - &- & - & - & -& -\\
Gu et al. \cite{gu2024efficient_cvpr2024}&Gen& CVPR/2024 & - & - & -& - & - & - & -& -& - & - & -& - &- & 44.3 & 58.6 & -& -\\
D4M \cite{su2024d_cvpr2024} &Gen& CVPR/2024 & - & 56.2 & 72.8& - & - & 45.0 & 48.8& -& - & - & 51.0 & 55.3 & -&34.2 & 63.4 & 66.5 & 68.1\\
IGD \cite{chen2025igd_iclr2025}&Gen& ICLR/2025 & - & - & 66.8& - & - & 45.8 & 53.9& 55.9& - & - & - & -& - &46.2 & 60.3 & -& -\\
\hline
SRe2L {\cite{yin2024squeeze_nips2024}}&SRe2L/R18& NeurIPS/2024 & 16.6 & 29.3 & 45.0& - & 6.6 & 27.0 & 50.2& -& 2.62&16.1 & 41.1 & 49.7& 0.4  & 21.3 & 46.8 & 52.8& 65.9\\
EDC \cite{shao2024elucidating_nips2024} &SRe2L/R18& NeurIPS/2024 & 32.6 & 79.1 & 87.0& - & 39.7 & 63.7 & 68.6& -&39.2 & 51.2 & 57.2& -& 12.8  & 48.6 & 58.0 & -& -\\
Xiao et al. \cite{xiao2024large_nips2024} &SRe2L/R18& NeurIPS/2024 & - & - & -& - & - & - & - & -&- & - & 48.8& 53.6& -  & 34.6 & 55.4 & 59.4 &62.6\\ 
RDED \cite{sun2024diversity_cvpr2024} &SRe2L/R18& CVPR/2024 & 22.9 & 37.1 & 62.1& - & 11.0 & 42.6 & 62.6& -&9.7 & 41.9 & 58.2& -& 6.6  & 42.0 & 56.5 & -& -\\
G-VBSM \cite{shao2024generalized_cvpr2024} &SRe2L/R18& CVPR/2024 & - & 53.5 & 59.2& - & 25.9 & 59.5 & 65.0& -& -&- & 47.6 & -& -  & 31.4 & 51.8 & 55.7& -\\ 
CDA \cite{yin2023dataset_arxiv2023} &SRe2L& TMLR/2024 & - & - & -& - & - & 49.8 & 64.4& -& -&21.3 & 48.7 & 53.2& 0.5  & 33.5 & 53.5 & 58.0& 63.3\\ 
SC-DD \cite{zhou2024self_arxiv2024} &SRe2L/R18& arXiv/2024 & - & - & -& - & - & - & 53.4& -& -&31.6 & 45.9 & -& -  & 32.1 & 53.1 & 57.9& 63.5\\ 
Zhong et al. \cite{zhong2024efficientdatasetdistillationdiffusiondriven_arxiv2024} &SRe2L/R18& arXiv/2024 & - & 36.4 & 61.0& - & - & 41.5 & 63.8 & -&- & 40.2 & 58.5& -& -  & 42.1 & 59.4 & 61.8& -\\
GIFT \cite{shang2024gift_arxiv2024}&SRe2L/R18& arXiv/2024 & - & - & -& - & - & 49.5 & 57.0 & -&- & 42.9 &47.5& -& -  & 21.7 & 39.5 & 42.5& -\\
CV-DD \cite{cui2025datasetdistillationcommitteevoting_arxiv2025} &SRe2L/R18& arXiv/2025 & - & 64.1 & 74.0& - & 28.3 & 62.7 & 67.1& -&10.1 & 47.8 & 54.1& -& -  & 46.0 & 59.5 & -& -\\
FocusDD \cite{hu2025focusddrealworldsceneinfusion_arxiv2025} &SRe2L/R18& arXiv/2025 & - & - & -& - & - & - & - & -&- & - & -& -& 8.8  & 45.3 & 61.7 & 62.0& -\\
\hline
DREAM \cite{liu2023dream_iccv2023} &Selective& ICCV/2023 &  51.1& 69.4 & 74.8& - &29.5 & 46.8 & 52.6 & -&10.0 & 29.5 & -& -& -  & 18.5 & - & -& -\\ 
DREAM+ \cite{liu2023dreamplus_arxiv2023} &Selective& arXiv/2023 &  52.5& 69.9 & 75.3& - &29.7 & 47.4 & 52.6 & -&10.5 & 24.0 & 29.5& -& -  & 18.5 & - & -& -\\ 
APM \cite{chen2023dataset_arxiv2023} &Selective& arXiv/2023 &  -& - & 75.0& - &-& 44.6 & 53.3 & 55.2&- & 30.0 & 38.2& 39.6& -  & 24.8 & 30.7 & 32.6&-\\
RFAD \cite{tukan2023dataset_arxiv2023} &Selective& arXiv/2023 &  64.4& 74.3 & 77.0& - &38.5& 45.8 & - & -&- & - & -& -& -  & - & - & -& -\\
PAD \cite{li2024prioritize_arxiv2024} &Selective& arXiv/2024 & 47.2 & 67.4 & 77.0& - & 28.4 & 47.8 & 55.9 & 58.5&17.7 & 32.3 & 41.6& -& -  & - & - & -& -\\ 
SDC \cite{wang2024sdc_arxiv2024} &Selective& arXiv/2024 &  47.9& 65.3 & 71.8& - &28.0& 47.8 & 52.5 & -&17.4 & 30.7 & 39.9& -& -  & - & - & -& -\\ 
AutoPalette \cite{yuancolor_nips2024} &Selective& NeurIPS/2024 & 58.6 & 74.3 & 79.4& - & 38.0 & 52.6 & 53.3 & -&- & - & -& -& -  & - & - & -& -\\ 
IADD \cite{li2024importance_nn} &Selective& NN/2024 &  46.5& 66.7 & 72.6& - &25.2 & 42.7 & 49.0 & -&9.6 & 24.1 & -& -& -  & - & - & -&-\\ 
BiLP+IDC \cite{xu2023distill_eccv2024}&Selective& ECCV/2024 &  55.9& 69.8 & 76.9& - &34.0& 48.0 & - & -&- & - & -& -& -  & - & - & -& -\\ 
SelMatch \cite{lee2024selmatch_icml2024} &Selective& ICML/2024 &  -& - & -& - &-& - & 54.5 & 62.4&- & - & 44.7& 50.4& -  & - & - & -&-\\
NRR-DD \cite{tran2025enhancingdatasetdistillationnoncritical_cvpr2025} &Selective& CVPR/2025 &  48.4& 66.7 & 73.1& - &27.3& 55.7 & 61.1 & -&20.4 & 44.3 & 50.2& -& 11.2  & 25.6 & 42.1 & -&-\\
CCFS\cite{chen2025curriculumcoarsetofineselectionhighipc_cvpr2025} &Selective& CVPR/2025 &  -& - & -& - &-& - & -& -& - & - & 55.8& 60.2& -& -& - & -&-\\
\hline
DATM \cite{guo2024towards_iclr2024} &Lossless& ICLR/2024 & 46.9 & 66.8 & 76.1& - & 27.9 & 47.2 & 55.0 & 57.5&17.1 & 31.1 & 39.7& -& -  & - & - & -& -\\ 
\hline
CUDD \cite{ma2024curriculum_arxiv2024} &Diversity/R18& arXiv/2024 & -& 56.2& 84.5 &  - &-& 60.3 & 65.7 & -&- & - & 55.6& 56.8& -  & 39.0 & 57.4 & 61.3 &65.0\\
DWA \cite{Du2024DiversityDrivenSE_nips2024} &Diversity/R18& NeurIPS/2024 &  -& 32.6 & 53.1& - &-& 39.6 & 60.9 & -&- & - & 52.8& 56.0& -  & 37.9 & 55.2 & 59.2 & -\\ 
INFER \cite{zhang2024breaking_arxiv2024} &Diversity/R18& ICLR/2025 &  -& 33.5 & 59.2& - &-& 50.2 & 65.1 & 68.6&- & 38.6 & 55.0& -& -  & 37.0 & 54.3 & -& -\\
DELT  \cite{shen2024delt_arxiv2024}  &Diversity/R18& CVPR/2025 &  24.0& 43.0 & 64.9& - &-& - & - & -&9.3 & 43.0 & 55.7& -& -  & 45.8 & 59.2 & 62.4&-\\
\hline
Zhang et al. \cite{zhang2023accelerating_cvpr2023} &Augmentation &CVPR/2023 &  49.2& 67.1 & 73.8& - &29.8& 45.6 & 52.6 & -&- & - & -& -& -  & - & - & - & -\\
LADD \cite{Kang2024LabelAugmentedDD_arxiv2024}&Augmentation/R18 &arXiv/2023 &  -& - & -& - &-& - & - & -&- & - & -& -& -  & 28.8 & - & - & -\\

        \hline
    \end{tabular}
\end{table*}

\begin{table}
\centering
\caption{Performance comparison of dataset distillation methods on ImageNet-21K under different IPCs.}
  \tiny
\begin{tabular}{|c|c|c|cccc|}
\hline
\multirow{2}{*}{\textbf{Methods}} & \multirow{2}{*}{\textbf{Schemes}} & \multirow{2}{*}{\textbf{Venue}} & \multicolumn{4}{c|}{\textbf{ImageNet-21K}} \\
\cline{4-7}
& & & 1 & 10 & 20 & 50 \\
\hline
SRe2L {\cite{yin2024squeeze_nips2024}}&SRe2L/R18& NeurIPS/2024 & - & 18.5& 21.8 & -\\
EDC \cite{shao2024elucidating_nips2024} &SRe2L/R18& NeurIPS/2024&- & 26.8& - & - \\
CDA \cite{yin2023dataset_arxiv2023} &SRe2L& TMLR/2024 & - & 22.6& 26.4 & - \\
RDED \cite{sun2024diversity_cvpr2024} &SRe2L/R18& CVPR/2024&- & 25.6& - & - \\
CUDD \cite{ma2024curriculum_arxiv2024} &Diversity/R18& arXiv/2024&- & 28.0& 34.9 & -\\
\hline
\end{tabular}
\label{tab:single-dataset-comparison}
\end{table}

\subsection{Performance and Scalability Across Techniques}
Different methodological approaches exhibit varying levels of success depending on dataset scale and IPC settings.
Early methods primarily focused on matching-based approaches, including gradient matching (GM) and trajectory matching (TM). 
While these methods demonstrated strong performance on smaller datasets (e.g., DC \cite{zhao2020dataset_iclr2021} and DSA \cite{zhao2021dataset_icmlr2021} achieving over 50\% accuracy on CIFAR-10), their effectiveness diminishes on larger-scale datasets like ImageNet-1K and ImageNet-21K. 
This scalability limitation stems from their difficulty in handling the increased complexity and high dimensionality of larger datasets.

Recent advances have shown that methods integrating SRe2L frameworks with diversity-enhancing strategies achieve superior performance on large-scale datasets. 
For instance, on ImageNet-1K, CUDD \cite{ma2024curriculum_arxiv2024} achieves 65.0\% accuracy at IPC=200, while D4M \cite{su2024d_cvpr2024} reaches 68.1\% under the same setting. 
These methods effectively address the complexity of larger datasets through their de-coupled optimization mechanisms and robust regularization strategies. 
Furthermore, their success extends to extremely large-scale scenarios, with methods like CUDD achieving 34.9\% accuracy on ImageNet-21K at IPC=20, demonstrating unprecedented scalability.

Emerging paradigms in dataset distillation include latent and frequency-based methods, such as FreD \cite{shin2024frequency_nips2023} and NSD \cite{yang2024neural_eccv2024}. 
These approaches show promising results on smaller datasets, with FreD achieving 60.6\% and NSD reaching 68.5\% accuracy on CIFAR-10 at IPC=1, significantly outperforming traditional matching-based methods. 
On CIFAR-100, they maintain strong performance with FreD achieving 34.6\% and NSD reaching 36.5\% at IPC=1. 
However, their scalability to larger datasets remains largely unexplored, as comprehensive results on ImageNet-1K and ImageNet-21K are yet to be reported.

Recent methods have also explored importance-based sampling and diversity-driven approaches. 
Notable examples include AutoPalette \cite{yuancolor_nips2024}, which achieves 58.6\% accuracy on CIFAR-10 at IPC=1, and EDC \cite{shao2024elucidating_nips2024}, which demonstrates strong performance across different scales, reaching 87.0\% on CIFAR-10 at IPC=50. 
These innovations suggest that combining multiple strategies, such as diversity enhancement, importance sampling, and efficient optimization, may be key for both scalability and performance.
In summary, while traditional matching-based methods remain effective for smaller datasets, the field has evolved towards more sophisticated approaches that combine de-coupling mechanisms like SRe2L, diversity-enhancing strategies, and novel paradigms such as latent and frequency-based methods. 
To advance the scalability and robustness of dataset distillation, future research must continue to address the challenges of large-scale datasets, particularly focusing on methods that can maintain high performance while scaling to datasets like ImageNet-21K.

\section{Challenges and Future Directions}

Despite significant advancements in dataset distillation, several fundamental challenges persist, particularly as the field scales to larger datasets, diverse architectures, and multimodal applications. Addressing these challenges requires a combination of theoretical insights, scalable algorithms, and robust evaluation frameworks.

\subsection{Challenges}

\noindent \textbf{Scalability to Large-Scale Datasets}  
While recent dataset distillation methods have demonstrated notable progress on ImageNet-1K, further scaling to ultra-large datasets such as ImageNet-21K remains a major challenge. 
Even state-of-the-art methods like CUDD \cite{ma2024curriculum_arxiv2024} achieve only 34.9\% accuracy at IPC=20, revealing a substantial performance drop when applied to datasets with extensive class diversity and intra-class variability. 
Computational constraints, increased dataset redundancy, and high memory costs further complicate large-scale distillation.

\noindent \textbf{Balancing Compression and Performance}  
The trade-off between dataset compression and performance varies significantly across dataset sizes and IPC levels. 
While significant performance improvements occur at lower IPC values (e.g., DATM \cite{guo2024towards_iclr2024} 
 improves from 46.9\% to 66.8\% on CIFAR-10 from IPC=1 to IPC=10), these gains diminish at higher IPCs. 
On CIFAR-100, PAD \cite{li2024prioritize_arxiv2024} shows only minor improvements from IPC=50 (55.9\%) to IPC=100 (58.5\%), suggesting a saturation point in dataset efficiency.

\noindent \textbf{Cross-Architecture Generalization} 
Current methods exhibit strong performance on pre-selected architectures but often fail to generalize across different model families. 
Vision Transformers  and emerging architectures demonstrate different feature learning mechanisms from traditional CNNs, making it unclear whether existing distillation techniques transfer effectively.

\noindent \textbf{Underexplored Domains and Modalities}  
Most existing distillation methods are designed for vision tasks, particularly image classification, while other core vision domains, such as object detection and semantic segmentation, remain largely underexplored \cite{Qi_2025_CVPR_2025}. 
Beyond vision, distillation in other modalities such as audio~\cite{jiang2024ddfad_arxiv2024}, 3D point clouds~\cite{yim2025permutation_arxiv2025}, and multimodal learning~\cite{xu2024low_arxiv2024} is still in its infancy. 
Challenges such as temporal consistency in video, frequency-space representations in audio, and cross-modal alignment in multimodal tasks are yet to be adequately addressed.

\noindent \textbf{Standardizing Fair Evaluation Practices}  
Recent methods increasingly rely on soft labels and teacher-student distillation strategies, making it difficult to separate the impact of distilled data from external knowledge transfer. 
Additionally, the use of varied evaluation protocols, loss functions, and augmentation strategies leads to inconsistencies in benchmarking, making cross-method comparisons less reliable.

\subsection{Future Directions}

\noindent \textbf{Ultra-Scalable Dataset Distillation}  
Achieving effective dataset distillation on billion-scale datasets like JFT-300M \cite{sun2017revisiting_iccv017} and LAION-5B \cite{schuhmann2022laion_nips2022} requires new strategies. 
Techniques such as dataset pruning integrated with distillation, and self-supervised data compression hold promise for tackling scalability challenges while maintaining high downstream task performance.

\noindent \textbf{Dynamic and Task-Aware Compression}  
Instead of relying on fixed IPC values, future methods should explore dynamic IPC optimization tailored to dataset complexity and target tasks. 
Meta-learning and reinforcement learning frameworks could be leveraged to automatically tune IPC levels based on data redundancy and model requirements.

\noindent \textbf{Towards Architecture-Agnostic Dataset Distillation}  
Developing distillation methods that generalize across architectures is crucial for ensuring broad applicability.
One promising direction is to distill representations that are invariant to specific network architectures. 
For example, future research could explore contrastive learning to enforce architecture-invariant feature alignment between CNNs and Transformers, ensuring that distilled datasets retain transferable representations.
Another potential avenue is developing adversarial strategies to enforce feature robustness across different model families.

\noindent \textbf{Towards Multimodal-Aware Dataset Distillation}  
Future directions in dataset distillation should move beyond conventional image classification benchmarks and address more complex and multimodal domains, such as medical imaging, time-series forecasting, and multimodal sensor fusion.
In these areas, it is essential to develop data representations tailored to specific modalities, such as spectral embeddings for audio signals or geometric descriptors for 3D point clouds \cite{yim2025permutation_arxiv2025}.
Importantly, even when the final distilled data set remains unimodal (e.g. image only), recent studies \cite{li2025leveragingmultimodalinformationenhance_arxiv2025} suggest that incorporating auxiliary multimodal knowledge, such as textual descriptions, segmentation masks, or audio cues, during the distillation process can significantly enhance the semantic richness and generalization ability of synthetic samples.
Using multimodal supervision during training can significantly improve distillation quality across both vision and non-vision domains.

\noindent \textbf{Robust and Fair Evaluation Frameworks}  
To ensure fair benchmarking, future studies should decouple dataset informativeness from external enhancements like teacher distillation and data augmentation. 
Recent efforts such as DD-Ranking \cite{li2024ddranking} introduce standardized evaluation metrics, but additional research is needed to refine benchmarking protocols and establish principled evaluation guidelines.

\noindent \textbf{Strengthening Theoretical Foundations}  
A rigorous theoretical framework is crucial for understanding dataset distillation’s limitations and guiding future advancements. 
Key challenges include quantifying information loss using entropy-based methods to establish lower bounds on minimal dataset requirements and applying PAC-learning \cite{valiant1984theory} to derive generalization guarantees. 
Understanding the role of spectral representations in bridging architectural or distributional gaps remains critical. 
Recent work, such as UniDD’s spectral filtering framework \cite{bo2025understandingdatasetdistillationspectral_arxiv2025}, unifies existing methods into a spectral framework. 
Exploring optimal transport theory \cite{10740308_tpami2025} for analyzing distributional shifts and integrating insights from information theory and statistical learning could also lead to more principled solutions.
As dataset distillation continues to evolve, establishing a solid theoretical foundation will be essential for ensuring its scalability, generalization, and robustness across  scenarios.

\section{Conclusion} \label{sec:conclusion}
This survey provides a comprehensive overview of the rapid advancements in dataset distillation from 2023 to 2025, with a particular focus on scaling to large-scale datasets and emerging methodological innovations. 
The field has witnessed significant progress across multiple dimensions, from achieving unprecedented performance on ImageNet-scale datasets to expanding into new domains like video, audio, and multi-modal processing.

Several key trends have emerged during this period. 
First, the development of more sophisticated optimization strategies, such as SRe2L frameworks and diversity-driven approaches, has enabled effective distillation of large-scale datasets like ImageNet-1K and ImageNet-21K. 
Second, the introduction of soft labels and decoupling mechanisms has significantly improved both efficiency and performance. 
Third, the emergence of generative models, particularly diffusion-based approaches, has opened new possibilities for high-quality synthetic data generation.
Despite these advances, important challenges remain. 
The scalability to ultra-large datasets, the trade-off between compression and performance, and cross-architecture generalization continue to be active areas of research. 
Additionally, the extension to multi-modal and temporal domains presents both opportunities and challenges that require innovative solutions.
We hope that this comprehensive survey of recent dataset distillation advances will serve as a valuable resource for researchers and practitioners, providing insights into the latest progress, existing challenges, and promising future directions in this rapidly evolving field.
\section*{Acknowledgment}
The authors would like to express their gratitude to the contributors of the {Awesome-Dataset-Distillation} repository \cite{li2022awesome}, which has served as a valuable resource in compiling recent advancements.
\ifCLASSOPTIONcaptionsoff
  \newpage
\fi



%



\footnotesize
\bibliographystyle{IEEEtran}
\bibliography{mainref.bib}

\begin{thebibliography}{100}
\providecommand{\url}[1]{#1}
\csname url@samestyle\endcsname
\providecommand{\newblock}{\relax}
\providecommand{\bibinfo}[2]{#2}
\providecommand{\BIBentrySTDinterwordspacing}{\spaceskip=0pt\relax}
\providecommand{\BIBentryALTinterwordstretchfactor}{4}
\providecommand{\BIBentryALTinterwordspacing}{\spaceskip=\fontdimen2\font plus
\BIBentryALTinterwordstretchfactor\fontdimen3\font minus \fontdimen4\font\relax}
\providecommand{\BIBforeignlanguage}[2]{{%
\expandafter\ifx\csname l@#1\endcsname\relax
\typeout{** WARNING: IEEEtran.bst: No hyphenation pattern has been}%
\typeout{** loaded for the language `#1'. Using the pattern for}%
\typeout{** the default language instead.}%
\else
\language=\csname l@#1\endcsname
\fi
#2}}
\providecommand{\BIBdecl}{\relax}
\BIBdecl

\bibitem{minaee2024large_arxiv2024}
S.~Minaee, T.~Mikolov, N.~Nikzad, M.~Chenaghlu, R.~Socher, X.~Amatriain, and J.~Gao, ``Large language models: A survey,'' \emph{arXiv preprint arXiv:2402.06196}, 2024.

\bibitem{zhang2024vision_tpami2024}
J.~Zhang, J.~Huang, S.~Jin, and S.~Lu, ``Vision-language models for vision tasks: A survey,'' \emph{IEEE Transactions on Pattern Analysis and Machine Intelligence}, 2024.

\bibitem{radford2021learning_icml2021}
A.~Radford, J.~W. Kim, C.~Hallacy, A.~Ramesh, G.~Goh, S.~Agarwal, G.~Sastry, A.~Askell, P.~Mishkin, J.~Clark \emph{et~al.}, ``Learning transferable visual models from natural language supervision,'' in \emph{ICML}, 2021.

\bibitem{wang2018dataset_arxiv2018}
T.~Wang, J.-Y. Zhu, A.~Torralba, and A.~A. Efros, ``Dataset distillation,'' \emph{arXiv preprint arXiv:1811.10959}, 2018.

\bibitem{yang2024dataset_icml2024}
W.~Yang, Y.~Zhu, Z.~Deng, and O.~Russakovsky, ``What is dataset distillation learning?'' in \emph{ICML}, 2024.

\bibitem{russakovsky2015imagenet}
O.~Russakovsky, J.~Deng, H.~Su, J.~Krause, S.~Satheesh, S.~Ma, Z.~Huang, A.~Karpathy, A.~Khosla, M.~Bernstein \emph{et~al.}, ``Imagenet large scale visual recognition challenge,'' \emph{International Journal of Computer Vision}, 2015.

\bibitem{ridnik2021imagenet_nips2021}
T.~Ridnik, E.~Ben-Baruch, A.~Noy, and L.~Zelnik-Manor, ``Imagenet-21k pretraining for the masses,'' in \emph{NeurIPS Datasets and Benchmarks}, 2021.

\bibitem{geng2023survey_ijcai2023}
J.~Geng, Z.~Chen, Y.~Wang, H.~Woisetschlaeger, S.~Schimmler, R.~Mayer, Z.~Zhao, and C.~Rong, ``A survey on dataset distillation: Approaches, applications and future directions,'' in \emph{IJCAI}, 2023.

\bibitem{lei2023comprehensive_tpami2023}
S.~Lei and D.~Tao, ``A comprehensive survey of dataset distillation,'' \emph{IEEE Transactions on Pattern Analysis and Machine Intelligence}, 2023.

\bibitem{yu2023dataset_tpami2023}
R.~Yu, S.~Liu, and X.~Wang, ``Dataset distillation: A comprehensive review,'' \emph{IEEE Transactions on Pattern Analysis and Machine Intelligence}, 2023.

\bibitem{deng2012mnist_spm2012}
L.~Deng, ``The mnist database of handwritten digit images for machine learning research [best of the web],'' \emph{IEEE signal processing magazine}, 2012.

\bibitem{werbos1990backpropagation}
P.~J. Werbos, ``Backpropagation through time: what it does and how to do it,'' \emph{Proceedings of the IEEE}, 1990.

\bibitem{feng2023embarrassingly_iclr2023}
Y.~Feng, S.~R. Vedantam, and J.~Kempe, ``Embarrassingly simple dataset distillation,'' in \emph{ICLR}, 2023.

\bibitem{yu2025teddy_eccv2024}
R.~Yu, S.~Liu, J.~Ye, and X.~Wang, ``Teddy: Efficient large-scale dataset distillation via taylor-approximated matching,'' in \emph{ECCV}, 2024.

\bibitem{zhao2020dataset_iclr2021}
B.~Zhao, K.~R. Mopuri, and H.~Bilen, ``Dataset condensation with gradient matching,'' in \emph{ICLR}, 2021.

\bibitem{zhao2021dataset_icmlr2021}
B.~Zhao and H.~Bilen, ``Dataset condensation with differentiable siamese augmentation,'' in \emph{ICML}, 2021.

\bibitem{cazenavette2022dataset_cvpr2022}
G.~Cazenavette, T.~Wang, A.~Torralba, A.~A. Efros, and J.-Y. Zhu, ``Dataset distillation by matching training trajectories,'' in \emph{CVPR}, 2022.

\bibitem{du2023minimizing_cvpr2023}
J.~Du, Y.~Jiang, V.~Y. Tan, J.~T. Zhou, and H.~Li, ``Minimizing the accumulated trajectory error to improve dataset distillation,'' in \emph{CVPR}, 2023.

\bibitem{krizhevsky2010convolutional}
A.~Krizhevsky, G.~Hinton \emph{et~al.}, ``Convolutional deep belief networks on cifar-10,'' \emph{Unpublished manuscript}, 2010.

\bibitem{shen2023ast_arxiv2023}
J.~Shen, W.~Yang, and K.-Y. Lam, ``Ast: Effective dataset distillation through alignment with smooth and high-quality expert trajectories,'' \emph{arXiv preprint arXiv:2310.10541}, 2023.

\bibitem{zhong2024towards_arxiv2024}
W.~Zhong, H.~Tang, Q.~Zheng, M.~Xu, Y.~Hu, and L.~Nie, ``Towards stable and storage-efficient dataset distillation: Matching convexified trajectory,'' in \emph{CVPR}, 2025.

\bibitem{jacot2018neural_nips2018}
A.~Jacot, F.~Gabriel, and C.~Hongler, ``Neural tangent kernel: Convergence and generalization in neural networks,'' in \emph{NeurIPS}, 2018.

\bibitem{li2024dataset_ieice2024}
G.~Li, R.~Togo, T.~Ogawa, and M.~Haseyama, ``Dataset distillation using parameter pruning,'' \emph{IEICE Transactions on Fundamentals of Electronics, Communications and Computer Sciences}, 2024.

\bibitem{cui2023scaling_icml2023}
J.~Cui, R.~Wang, S.~Si, and C.-J. Hsieh, ``Scaling up dataset distillation to imagenet-1k with constant memory,'' in \emph{ICML}, 2023.

\bibitem{liu2024dataset_eccv2024}
D.~Liu, J.~Gu, H.~Cao, C.~Trinitis, and M.~Schulz, ``Dataset distillation by automatic training trajectories,'' in \emph{ECCV}, 2024.

\bibitem{kong2025efficient_icassp2025}
H.~Kong, W.~Zhou, X.~He, X.~Tu, and X.~Ding, ``Efficient dataset distillation through low-rank space sampling,'' in \emph{ICASSP}, 2025.

\bibitem{zhao2023dataset_wacv2023}
B.~Zhao and H.~Bilen, ``Dataset condensation with distribution matching,'' in \emph{WACV}, 2023.

\bibitem{wang2022cafe_cvpr2022}
K.~Wang, B.~Zhao, X.~Peng, Z.~Zhu, S.~Yang, S.~Wang, G.~Huang, H.~Bilen, X.~Wang, and Y.~You, ``Cafe: Learning to condense dataset by aligning features,'' in \emph{CVPR}, 2022.

\bibitem{zhao2023improved_cvpr2023}
G.~Zhao, G.~Li, Y.~Qin, and Y.~Yu, ``Improved distribution matching for dataset condensation,'' in \emph{CVPR}, 2023.

\bibitem{Zhang2024_ijcai2024}
H.~Zhang, S.~Li, F.~Lin, W.~Wang, Z.~Qian, and S.~Ge, ``Dance: Dual-view distribution alignment for dataset condensation,'' in \emph{IJCAI}, 2024.

\bibitem{Malakshan2024distilling_arxiv2024}
S.~R. Malakshan, M.~S.~E. Saadabadi, A.~Dabouei, and N.~M. Nasrabadi, ``Decomposed distribution matching in dataset condensation,'' in \emph{WACV}, 2025.

\bibitem{sajedi2023datadam_iccv2023}
A.~Sajedi, S.~Khaki, E.~Amjadian, L.~Z. Liu, Y.~A. Lawryshyn, and K.~N. Plataniotis, ``Datadam: Efficient dataset distillation with attention matching,'' in \emph{ICCV}, 2023.

\bibitem{zhang2024m3d_aaai2024}
H.~Zhang, S.~Li, P.~Wang, D.~Zeng, and S.~Ge, ``M3d: Dataset condensation by minimizing maximum mean discrepancy,'' in \emph{AAAI}, 2024.

\bibitem{berlinet2011reproducing}
A.~Berlinet and C.~Thomas-Agnan, \emph{Reproducing kernel Hilbert spaces in probability and statistics}.\hskip 1em plus 0.5em minus 0.4em\relax Springer Science \& Business Media, 2011.

\bibitem{wei2024dataset_cvprw2024}
W.~Wei, T.~De~Schepper, and K.~Mets, ``Dataset condensation with latent quantile matching,'' in \emph{CVPR Workshop}, 2024.

\bibitem{liu2023dataset_arxiv2023}
H.~Liu, Y.~Li, T.~Xing, V.~Dalal, L.~Li, J.~He, and H.~Wang, ``Dataset distillation via the wasserstein metric,'' \emph{arXiv preprint arXiv:2311.18531}, 2023.

\bibitem{panaretos2019statistical}
V.~M. Panaretos and Y.~Zemel, ``Statistical aspects of wasserstein distances,'' \emph{Annual review of statistics and its application}, 2019.

\bibitem{deng2024exploiting_cvpr2024}
W.~Deng, W.~Li, T.~Ding, L.~Wang, H.~Zhang, K.~Huang, J.~Huo, and Y.~Gao, ``Exploiting inter-sample and inter-feature relations in dataset distillation,'' in \emph{CVPR}, 2024.

\bibitem{wang2025dataset_cvpr2025}
S.~Wang, Y.~Yang, Z.~Liu, C.~Sun, X.~Hu, C.~He, and L.~Zhang, ``Dataset distillation with neural characteristic function: A minmax perspective,'' in \emph{CVPR}, 2025.

\bibitem{li2025hyperbolicdatasetdistillation_arxiv2025}
W.~Li, G.~Li, K.~Maeda, T.~Ogawa, and M.~Haseyama, ``Hyperbolic dataset distillation,'' \emph{arXiv preprint arXiv:2505.24623}, 2025.

\bibitem{duan2023dataset_arxiv2023}
Y.~Duan, J.~Zhang, and L.~Zhang, ``Dataset distillation in latent space,'' \emph{arXiv preprint arXiv:2311.15547}, 2023.

\bibitem{cazenavette2023generalizing_cvpr2023}
G.~Cazenavette, T.~Wang, A.~Torralba, A.~A. Efros, and J.-Y. Zhu, ``Generalizing dataset distillation via deep generative prior,'' in \emph{CVPR}, 2023.

\bibitem{sauer2022stylegan_acmsiggraph}
A.~Sauer, K.~Schwarz, and A.~Geiger, ``Stylegan-xl: Scaling stylegan to large diverse datasets,'' in \emph{ACM SIGGRAPH}, 2022.

\bibitem{li2025datasetdistillationprobabilisticlatent_arxiv2025}
Z.~Li, S.~Cechnicka, C.~Ouyang, K.~Breininger, P.~Schüffler, and B.~Kainz, ``Dataset distillation with probabilistic latent features,'' \emph{arXiv preprint arXiv:2505.06647}, 2025.

\bibitem{shin2024frequency_nips2023}
D.~Shin, S.~Shin, and I.-C. Moon, ``Frequency domain-based dataset distillation,'' in \emph{NeurIPS}, 2023.

\bibitem{yang2024neural_eccv2024}
S.~Yang, S.~Cheng, M.~Hong, H.~Fan, X.~Wei, and S.~Liu, ``Neural spectral decomposition for dataset distillation,'' in \emph{ECCV}, 2024.

\bibitem{shin2025distilling_iclr2025}
D.~Shin, H.~Bae, G.~Sim, W.~Kang, and I.~chul Moon, ``Distilling dataset into neural field,'' in \emph{ICLR}, 2025.

\bibitem{bo2025understandingdatasetdistillationspectral_arxiv2025}
D.~Bo, S.~Liu, and X.~Wang, ``Understanding dataset distillation via spectral filtering,'' \emph{arXiv preprint arXiv:2503.01212}, 2025.

\bibitem{du2024sequential_nips2024}
J.~Du, Q.~Shi, and J.~T. Zhou, ``Sequential subset matching for dataset distillation,'' in \emph{NeurIPS}, 2023.

\bibitem{cui2025optical_cvpr2025}
X.~Cui, Y.~Qin, W.~Zhou, H.~Li, and H.~Li, ``Optical: Leveraging optimal transport for contribution allocation in dataset distillation,'' in \emph{CVPR}, 2025.

\bibitem{belghazi2018mutual_icml2018}
M.~I. Belghazi, A.~Baratin, S.~Rajeshwar, S.~Ozair, Y.~Bengio, A.~Courville, and D.~Hjelm, ``Mutual information neural estimation,'' in \emph{ICML}, 2018.

\bibitem{shang2024mim4dd_nips2023}
Y.~Shang, Z.~Yuan, and Y.~Yan, ``Mim4dd: Mutual information maximization for dataset distillation,'' in \emph{NeurIPS}, 2023.

\bibitem{zhong2024going_arxiv2024}
X.~Zhong, B.~Chen, H.~Fang, X.~Gu, S.-T. Xia, and E.-H. Yang, ``Going beyond feature similarity: Effective dataset distillation based on class-aware conditional mutual information,'' in \emph{ICLR}, 2025.

\bibitem{son2024fyi_eccv2024}
B.~Son, Y.~Oh, D.~Baek, and B.~Ham, ``Fyi: Flip your images for dataset distillation,'' in \emph{ECCV}, 2024.

\bibitem{lu2023can_arxiv2023}
Y.~Lu, X.~Chen, Y.~Zhang, J.~Gu, T.~Zhang, Y.~Zhang, X.~Yang, Q.~Xuan, K.~Wang, and Y.~You, ``Can pre-trained models assist in dataset distillation?'' \emph{arXiv preprint arXiv:2310.03295}, 2023.

\bibitem{zhong2024hierarchical_cvpr2025}
X.~Zhong, H.~Fang, B.~Chen, X.~Gu, T.~Dai, M.~Qiu, and S.-T. Xia, ``Hierarchical features matter: A deep exploration of gan priors for improved dataset distillation,'' in \emph{CVPR}, 2025.

\bibitem{khosla2020supervised_nips2020}
P.~Khosla, P.~Teterwak, C.~Wang, A.~Sarna, Y.~Tian, P.~Isola, A.~Maschinot, C.~Liu, and D.~Krishnan, ``Supervised contrastive learning,'' in \emph{NeurIPS}, 2020.

\bibitem{torkkola2003feature_jmlr2003}
K.~Torkkola, ``Feature extraction by non-parametric mutual information maximization,'' \emph{Journal of Machine Learning Research}, 2003.

\bibitem{goodfellow2014generative}
I.~Goodfellow, J.~Pouget-Abadie, M.~Mirza, B.~Xu, D.~Warde-Farley, S.~Ozair, A.~Courville, and Y.~Bengio, ``Generative adversarial nets,'' in \emph{NeurIPS}, 2014.

\bibitem{croitoru2023diffusion}
F.-A. Croitoru, V.~Hondru, R.~T. Ionescu, and M.~Shah, ``Diffusion models in vision: A survey,'' \emph{IEEE Transactions on Pattern Analysis and Machine Intelligence}, 2023.

\bibitem{yin2024squeeze_nips2024}
Z.~Yin, E.~Xing, and Z.~Shen, ``Squeeze, recover and relabel: Dataset condensation at imagenet scale from a new perspective,'' in \emph{NeurIPS}, 2024.

\bibitem{qin2024label_arxiv2024}
T.~Qin, Z.~Deng, and D.~Alvarez-Melis, ``A label is worth a thousand images in dataset distillation,'' in \emph{NeurIPS}, 2024.

\bibitem{zhao2022synthesizing_arxiv2022}
B.~Zhao and H.~Bilen, ``Synthesizing informative training samples with gan,'' in \emph{NeurIPS Workshop}, 2022.

\bibitem{li2024generative_cvpr2024}
L.~Li, G.~Li, R.~Togo, K.~Maeda, T.~Ogawa, and M.~Haseyama, ``Generative dataset distillation: Balancing global structure and local details,'' in \emph{CVPR Workshop}, 2024.

\bibitem{wang2023dim_arxiv2023}
K.~Wang, J.~Gu, D.~Zhou, Z.~Zhu, W.~Jiang, and Y.~You, ``Dim: Distilling dataset into generative model,'' in \emph{ECCV Workshop}, 2024.

\bibitem{zhang2023dataset_arxiv2023}
D.~J. Zhang, H.~Wang, C.~Xue, R.~Yan, W.~Zhang, S.~Bai, and M.~Z. Shou, ``Dataset condensation via generative model,'' \emph{arXiv preprint arXiv:2309.07698}, 2023.

\bibitem{gu2024efficient_cvpr2024}
J.~Gu, S.~Vahidian, V.~Kungurtsev, H.~Wang, W.~Jiang, Y.~You, and Y.~Chen, ``Efficient dataset distillation via minimax diffusion,'' in \emph{CVPR}, 2024.

\bibitem{su2024d_cvpr2024}
D.~Su, J.~Hou, W.~Gao, Y.~Tian, and B.~Tang, ``D\^{} 4: Dataset distillation via disentangled diffusion model,'' in \emph{CVPR}, 2024.

\bibitem{wang2025_iccv2025}
H.~Wang, Z.~Zhao, J.~Wu, Y.~Shang, G.~Liu, and Y.~Yan, ``Cao$_2$: Rectifying inconsistencies in diffusion-based dataset distillation,'' in \emph{ICCV}, 2025.

\bibitem{zou2025_iccv2025}
Y.~Zou, G.~Li, D.~Su, Z.~Wang, J.~Yu, and C.~Zhang, ``Dataset distillation via vision-language category prototype,'' in \emph{ICCV}, 2025.

\bibitem{abbasi2024one_arxiv2024}
A.~Abbasi, A.~Shahbazi, H.~Pirsiavash, and S.~Kolouri, ``One category one prompt: Dataset distillation using diffusion models,'' \emph{arXiv preprint arXiv:2403.07142}, 2024.

\bibitem{2024arXiv241204668A_arxiv2024}
A.~{Abbasi}, S.~{Imani}, C.~{An}, G.~{Mahalingam}, H.~{Shrivastava}, M.~{Diesendruck}, H.~{Pirsiavash}, P.~{Sharma}, and S.~{Kolouri}, ``{Diffusion-Augmented Coreset Expansion for Scalable Dataset Distillation},'' \emph{arXiv preprint arXiv:2412.04668}, 2024.

\bibitem{sun2024diversity_cvpr2024}
P.~Sun, B.~Shi, D.~Yu, and T.~Lin, ``On the diversity and realism of distilled dataset: An efficient dataset distillation paradigm,'' in \emph{CVPR}, 2024.

\bibitem{moser2024latent_arxiv2024}
B.~B. Moser, F.~Raue, S.~Palacio, S.~Frolov, and A.~Dengel, ``Latent dataset distillation with diffusion models,'' \emph{arXiv preprint arXiv:2403.03881}, 2024.

\bibitem{zhao2025taming_icml2025}
L.~Zhao, Y.~Wu, X.~Jiang, J.~Gu, Y.~Wang, X.~Xu, P.~Zhao, and X.~Lin, ``Taming diffusion for dataset distillation with high representativeness,'' in \emph{ICML}, 2025.

\bibitem{qin2024distributional_arxiv2024}
T.~Qin, Z.~Deng, and D.~Alvarez-Melis, ``Distributional dataset distillation with subtask decomposition,'' \emph{arXiv preprint arXiv:2403.00999}, 2024.

\bibitem{10571602_tpami2024}
W.~Huang, M.~Ye, Z.~Shi, G.~Wan, H.~Li, B.~Du, and Q.~Yang, ``Federated learning for generalization, robustness, fairness: A survey and benchmark,'' \emph{IEEE Transactions on Pattern Analysis and Machine Intelligence}, 2024.

\bibitem{chen2025igd_iclr2025}
M.~Chen, J.~Du, B.~Huang, Y.~Wang, X.~Zhang, and W.~Wang, ``Influence-guided diffusion for dataset distillation,'' in \emph{ICLR}, 2025.

\bibitem{peebles2023scalable_iccv2023}
W.~Peebles and S.~Xie, ``Scalable diffusion models with transformers,'' in \emph{ICCV}, 2023.

\bibitem{chan2025mgd_icml2025}
J.~A. Chan-Santiago, P.~Tirupattur, G.~K. Nayak, G.~Liu, and M.~Shah, ``Mgd3: Mode-guided dataset distillation using diffusion models,'' in \emph{ICML}, 2025.

\bibitem{lipman2022flow_arxiv2022}
Y.~Lipman, R.~T. Chen, H.~Ben-Hamu, M.~Nickel, and M.~Le, ``Flow matching for generative modeling,'' in \emph{ICLR}, 2023.

\bibitem{chen2024flow_iclr2024}
R.~T. Chen and Y.~Lipman, ``Flow matching on general geometries,'' in \emph{ICLR}, 2024.

\bibitem{cai2024batch_cvpr2024}
Z.~Cai, H.~Zhu, Q.~Shen, X.~Wang, and X.~Cao, ``Batch normalization alleviates the spectral bias in coordinate networks,'' in \emph{CVPR}, 2024.

\bibitem{yin2023dataset_arxiv2023}
Z.~Yin and Z.~Shen, ``Dataset distillation in large data era,'' \emph{Transactions on Machine Learning Research}, 2024.

\bibitem{jiang2015self_aaai2015}
L.~Jiang, D.~Meng, Q.~Zhao, S.~Shan, and A.~Hauptmann, ``Self-paced curriculum learning,'' in \emph{AAAI}, 2015.

\bibitem{zhou2024self_arxiv2024}
M.~Zhou, Z.~Yin, S.~Shao, and Z.~Shen, ``Self-supervised dataset distillation: A good compression is all you need,'' \emph{arXiv preprint arXiv:2404.07976}, 2024.

\bibitem{zong2024self_pami2024}
Y.~Zong, O.~Mac~Aodha, and T.~Hospedales, ``Self-supervised multimodal learning: A survey,'' \emph{IEEE Transactions on Pattern Analysis and Machine Intelligence}, 2024.

\bibitem{shao2024generalized_cvpr2024}
S.~Shao, Z.~Yin, M.~Zhou, X.~Zhang, and Z.~Shen, ``Generalized large-scale data condensation via various backbone and statistical matching,'' in \emph{CVPR}, 2024.

\bibitem{shao2024elucidating_nips2024}
S.~Shao, Z.~Zhou, H.~Chen, and Z.~Shen, ``Elucidating the design space of dataset condensation,'' in \emph{NeurIPS}, 2024.

\bibitem{cui2025datasetdistillationcommitteevoting_arxiv2025}
J.~Cui, Z.~Li, X.~Ma, X.~Bi, Y.~Luo, and Z.~Shen, ``Dataset distillation via committee voting,'' \emph{arXiv preprint arXiv:2501.07575}, 2025.

\bibitem{shen2022fast_eccv2022}
Z.~Shen and E.~Xing, ``A fast knowledge distillation framework for visual recognition,'' in \emph{ECCV}, 2022.

\bibitem{hu2025focusddrealworldsceneinfusion_arxiv2025}
Y.~Hu, Y.~Cheng, O.~Saukh, F.~Ozdemir, A.~Lu, Z.~Cao, and Z.~Li, ``Focusdd: Real-world scene infusion for robust dataset distillation,'' \emph{arXiv preprint arXiv:2501.06405}, 2025.

\bibitem{50650_iclr2021}
A.~Kolesnikov, A.~Dosovitskiy, D.~Weissenborn, G.~Heigold, J.~Uszkoreit, L.~Beyer, M.~Minderer, M.~Dehghani, N.~Houlsby, S.~Gelly, T.~Unterthiner, and X.~Zhai, ``An image is worth 16x16 words: Transformers for image recognition at scale,'' in \emph{ICLR}, 2021.

\bibitem{zhong2024efficientdatasetdistillationdiffusiondriven_arxiv2024}
X.~Zhong, S.~Sun, X.~Gu, Z.~Xu, Y.~Wang, J.~Wu, and B.~Chen, ``Efficient dataset distillation via diffusion-driven patch selection for improved generalization,'' \emph{arXiv preprint arXiv:2412.09959}, 2024.

\bibitem{shang2024gift_arxiv2024}
X.~Shang, P.~Sun, and T.~Lin, ``Gift: Unlocking full potential of labels in distilled dataset at near-zero cost,'' in \emph{ICLR}, 2025.

\bibitem{xiao2024large_nips2024}
L.~Xiao and Y.~He, ``Are large-scale soft labels necessary for large-scale dataset distillation?'' in \emph{NeurIPS}, 2024.

\bibitem{wang2024emphasizing_arxiv2024}
K.~Wang, Z.~Li, Z.-Q. Cheng, S.~Khaki, A.~Sajedi, R.~Vedantam, K.~N. Plataniotis, A.~Hauptmann, and Y.~You, ``Emphasizing discriminative features for dataset distillation in complex scenarios,'' in \emph{CVPR}, 2025.

\bibitem{selvaraju2017grad_iccv2017}
R.~R. Selvaraju, M.~Cogswell, A.~Das, R.~Vedantam, D.~Parikh, and D.~Batra, ``Grad-cam: Visual explanations from deep networks via gradient-based localization,'' in \emph{ICCV}, 2017.

\bibitem{tran2025enhancingdatasetdistillationnoncritical_cvpr2025}
M.-T. Tran, T.~Le, X.-M. Le, T.-T. Do, and D.~Phung, ``Enhancing dataset distillation via non-critical region refinement,'' in \emph{CVPR}, 2025.

\bibitem{yuancolor_nips2024}
B.~Yuan, Z.~Wang, M.~Baktashmotlagh, Y.~Luo, and Z.~Huang, ``Color-oriented redundancy reduction in dataset distillation,'' in \emph{NeurIPS}, 2024.

\bibitem{li2024importance_nn}
G.~Li, R.~Togo, T.~Ogawa, and M.~Haseyama, ``Importance-aware adaptive dataset distillation,'' \emph{Neural Networks}, 2024.

\bibitem{liu2023dream_iccv2023}
Y.~Liu, J.~Gu, K.~Wang, Z.~Zhu, W.~Jiang, and Y.~You, ``Dream: Efficient dataset distillation by representative matching,'' in \emph{ICCV}, 2023.

\bibitem{liu2023dreamplus_arxiv2023}
Y.~Liu, J.~Gu, K.~Wang, Z.~Zhu, K.~Zhang, W.~Jiang, and Y.~You, ``Dream+: Efficient dataset distillation by bidirectional representative matching,'' \emph{arXiv preprint arXiv:2310.15052}, 2023.

\bibitem{moser2024distill_arxiv2024}
B.~B. Moser, F.~Raue, T.~C. Nauen, S.~Frolov, and A.~Dengel, ``Distill the best, ignore the rest: Improving dataset distillation with loss-value-based pruning,'' \emph{arXiv preprint arXiv:2411.12115}, 2024.

\bibitem{xu2023distill_eccv2024}
Y.~Xu, Y.-L. Li, K.~Cui, Z.~Wang, C.~Lu, Y.-W. Tai, and C.-K. Tang, ``Distill gold from massive ores: Efficient dataset distillation via critical samples selection,'' in \emph{ECCV}, 2024.

\bibitem{wang2024sdc_arxiv2024}
S.~Wang, Y.~Yang, Q.~Wang, K.~Li, L.~Zhang, and Y.~Junchi, ``Not all samples should be utilized equally: Towards understanding and improving dataset distillation,'' in \emph{CVPR Workshop}, 2025.

\bibitem{lee2024selmatch_icml2024}
Y.~Lee and H.~W. Chung, ``Selmatch: Effectively scaling up dataset distillation via selection-based initialization and partial updates by trajectory matching,'' in \emph{ICML}, 2024.

\bibitem{chen2025curriculumcoarsetofineselectionhighipc_cvpr2025}
Y.~Chen, G.~Chen, M.~Zhang, W.~Guan, and L.~Nie, ``Curriculum coarse-to-fine selection for high-ipc dataset distillation,'' in \emph{CVPR}, 2025.

\bibitem{chen2023dataset_arxiv2023}
M.~Chen, B.~Huang, J.~Lu, B.~Li, Y.~Wang, M.~Cheng, and W.~Wang, ``Dataset distillation via adversarial prediction matching,'' \emph{arXiv preprint arXiv:2312.08912}, 2023.

\bibitem{tukan2023dataset_arxiv2023}
M.~Tukan, A.~Maalouf, and M.~Osadchy, ``Dataset distillation meets provable subset selection,'' \emph{arXiv preprint arXiv:2307.08086}, 2023.

\bibitem{li2024prioritize_arxiv2024}
Z.~Li, Z.~Guo, W.~Zhao, T.~Zhang, Z.-Q. Cheng, S.~Khaki, K.~Zhang, A.~Sajed, K.~N. Plataniotis, K.~Wang \emph{et~al.}, ``Prioritize alignment in dataset distillation,'' \emph{arXiv preprint arXiv:2408.03360}, 2024.

\bibitem{paul2021deep_nips2021}
M.~Paul, S.~Ganguli, and G.~K. Dziugaite, ``Deep learning on a data diet: Finding important examples early in training,'' in \emph{NeurIPS}, 2021.

\bibitem{guo2024towards_iclr2024}
Z.~Guo, K.~Wang, G.~Cazenavette, H.~Li, K.~Zhang, and Y.~You, ``Towards lossless dataset distillation via difficulty-aligned trajectory matching,'' in \emph{ICLR}, 2024.

\bibitem{zhou2024enhancing_arxiv2024}
C.~Zhou, C.~Jiang, Y.~Xie, H.~Cao, and J.~Yang, ``Enhancing dataset distillation via label inconsistency elimination and learning pattern refinement,'' in \emph{ECCV 2024 Dataset Distillation Challenge}, 2024.

\bibitem{ma2024curriculum_arxiv2024}
Z.~Ma, A.~Cao, F.~Yang, and X.~Wei, ``Curriculum dataset distillation,'' \emph{arXiv preprint arXiv:2405.09150}, 2024.

\bibitem{zhang2024breaking_arxiv2024}
X.~Zhang, J.~Du, P.~Liu, and J.~T. Zhou, ``Breaking class barriers: Efficient dataset distillation via inter-class feature compensator,'' in \emph{ICLR}, 2025.

\bibitem{shen2024delt_arxiv2024}
Z.~Shen, A.~Sherif, Z.~Yin, and S.~Shao, ``Delt: A simple diversity-driven earlylate training for dataset distillation,'' in \emph{CVPR}, 2025.

\bibitem{Du2024DiversityDrivenSE_nips2024}
J.~Du, X.~Zhang, J.~Hu, W.~Huang, and J.~T. Zhou, ``Diversity-driven synthesis: Enhancing dataset distillation through directed weight adjustment,'' in \emph{NeurIPS}, 2024.

\bibitem{li2024diversified_mm2024}
H.~Li, Y.~Zhou, X.~Gu, B.~Li, and W.~Wang, ``Diversified semantic distribution matching for dataset distillation,'' in \emph{ACM MM}, 2024.

\bibitem{zhang2023accelerating_cvpr2023}
L.~Zhang, J.~Zhang, B.~Lei, S.~Mukherjee, X.~Pan, B.~Zhao, C.~Ding, Y.~Li, and D.~Xu, ``Accelerating dataset distillation via model augmentation,'' in \emph{CVPR}, 2023.

\bibitem{Kang2024LabelAugmentedDD_arxiv2024}
S.~Kang, Y.~Lim, and H.~Shim, ``Label-augmented dataset distillation,'' in \emph{WACV}, 2025.

\bibitem{10172347_pami2023}
Z.~Zhu, K.~Lin, A.~K. Jain, and J.~Zhou, ``Transfer learning in deep reinforcement learning: A survey,'' \emph{IEEE Transactions on Pattern Analysis and Machine Intelligence}, 2023.

\bibitem{shul2024distilling_arxiv2024}
A.~Shul, E.~Horwitz, and Y.~Hoshen, ``Distilling datasets into less than one image,'' \emph{arXiv preprint arXiv:2403.12040}, 2024.

\bibitem{li2025contrastivelearningenhancedtrajectorymatching_arxiv2025}
W.~Li, S.~Sakai, and T.~Hasegawa, ``Contrastive learning-enhanced trajectory matching for small-scale dataset distillation,'' \emph{arXiv preprint arXiv:2505.15267}, 2025.

\bibitem{yang2024generalized_IJCV2024}
J.~Yang, K.~Zhou, Y.~Li, and Z.~Liu, ``Generalized out-of-distribution detection: A survey,'' \emph{International Journal of Computer Vision}, 2024.

\bibitem{jiang2022dataset_pami2022}
S.~Jiang, Y.~Zhu, C.~Liu, X.~Song, X.~Li, and W.~Min, ``Dataset bias in few-shot image recognition,'' \emph{IEEE transactions on Pattern Analysis and Machine Intelligence}, 2022.

\bibitem{huang2024federated_pami2024}
W.~Huang, M.~Ye, Z.~Shi, G.~Wan, H.~Li, B.~Du, and Q.~Yang, ``Federated learning for generalization, robustness, fairness: A survey and benchmark,'' \emph{IEEE Transactions on Pattern Analysis and Machine Intelligence}, 2024.

\bibitem{gui2024survey_tpami2024}
J.~Gui, T.~Chen, J.~Zhang, Q.~Cao, Z.~Sun, H.~Luo, and D.~Tao, ``A survey on self-supervised learning: Algorithms, applications, and future trends,'' \emph{IEEE Transactions on Pattern Analysis and Machine Intelligence}, 2024.

\bibitem{5288526_kdd2010}
S.~J. Pan and Q.~Yang, ``A survey on transfer learning,'' \emph{IEEE Transactions on Knowledge and Data Engineering}, 2010.

\bibitem{ma2025towards_pr2025}
S.~Ma, F.~Zhu, Z.~Cheng, and X.-Y. Zhang, ``Towards trustworthy dataset distillation,'' \emph{Pattern Recognition}, 2025.

\bibitem{choi2025damdomainawaremodulemultidomain_arxiv2025}
J.~Choi, G.~Han, D.-J. Lee, S.~Baek, and J.~Kim, ``Dam: Domain-aware module for multi-domain dataset condensation,'' \emph{arXiv preprint arXiv:2505.22387}, 2025.

\bibitem{lu2024exploring_cvpr2024}
Y.~Lu, J.~Gu, X.~Chen, S.~Vahidian, and Q.~Xuan, ``Exploring the impact of dataset bias on dataset distillation,'' in \emph{CVPR Workshop}, 2024.

\bibitem{cuimitigating_icml2024}
J.~Cui, R.~Wang, Y.~Xiong, and C.-J. Hsieh, ``Mitigating bias in dataset distillation,'' in \emph{ICML}, 2024.

\bibitem{zhou2024fairdd_arxiv2024}
Q.~Zhou, S.~Fang, S.~He, W.~Meng, and J.~Chen, ``Fairdd: Fair dataset distillation via synchronized matching,'' \emph{arXiv preprint arXiv:2411.19623}, 2024.

\bibitem{zhao2024distilling_arxiv2024}
Z.~Zhao, H.~Wang, Y.~Shang, K.~Wang, and Y.~Yan, ``Distilling long-tailed datasets,'' in \emph{CVPR}, 2025.

\bibitem{lee2023self_iclr2024}
D.~B. Lee, S.~Lee, J.~Ko, K.~Kawaguchi, J.~Lee, and S.~J. Hwang, ``Self-supervised set representation learning for unsupervised meta-learning,'' in \emph{ICLR}, 2024.

\bibitem{yu2025self_iclr2025}
S.-F. Yu, J.-J. Yao, and W.-C. Chiu, ``Boost self-supervised dataset distillation via parameterization, predefined augmentation, and approximation,'' in \emph{ICLR}, 2025.

\bibitem{joshi2024dataset_arxiv2024}
S.~Joshi, J.~Ni, and B.~Mirzasoleiman, ``Dataset distillation via knowledge distillation: Towards efficient self-supervised pre-training of deep networks,'' in \emph{ICLR}, 2025.

\bibitem{zhou2020distilled_arxiv2020}
Y.~Zhou, G.~Pu, X.~Ma, X.~Li, and D.~Wu, ``Distilled one-shot federated learning,'' \emph{arXiv preprint arXiv:2009.07999}, 2020.

\bibitem{hu2022fedsynth_arxiv2022}
S.~Hu, J.~Goetz, K.~Malik, H.~Zhan, Z.~Liu, and Y.~Liu, ``Fedsynth: Gradient compression via synthetic data in federated learning,'' in \emph{NeurIPS Workshop}, 2022.

\bibitem{zhang2022dense_nips2022}
J.~Zhang, C.~Chen, B.~Li, L.~Lyu, S.~Wu, S.~Ding, C.~Shen, and C.~Wu, ``Dense: Data-free one-shot federated learning,'' in \emph{NeurIPS}, 2022.

\bibitem{liu2022meta_iclr2023}
P.~Liu, X.~Yu, and J.~T. Zhou, ``Meta knowledge condensation for federated learning,'' in \emph{ICLR}, 2023.

\bibitem{song2023federated_ijcnn2023}
R.~Song, D.~Liu, D.~Z. Chen, A.~Festag, C.~Trinitis, M.~Schulz, and A.~Knoll, ``Federated learning via decentralized dataset distillation in resource-constrained edge environments,'' in \emph{IJCNN}, 2023.

\bibitem{pan2024fedcache_ap2024}
Q.~Pan, S.~Sun, Z.~Wu, Y.~Wang, M.~Liu, B.~Gao, and J.~Wang, ``Fedcache 2.0: Federated edge learning with knowledge caching and dataset distillation,'' \emph{Authorea Preprints}, 2024.

\bibitem{xiong2023feddm_cvpr2023}
Y.~Xiong, R.~Wang, M.~Cheng, F.~Yu, and C.-J. Hsieh, ``Feddm: Iterative distribution matching for communication-efficient federated learning,'' in \emph{CVPR}, 2023.

\bibitem{pi2023dynafed_cvpr2023}
R.~Pi, W.~Zhang, Y.~Xie, J.~Gao, X.~Wang, S.~Kim, and Q.~Chen, ``Dynafed: Tackling client data heterogeneity with global dynamics,'' in \emph{CVPR}, 2023.

\bibitem{wang2024aggregation_cvpr2024}
Y.~Wang, H.~Fu, R.~Kanagavelu, Q.~Wei, Y.~Liu, and R.~S.~M. Goh, ``An aggregation-free federated learning for tackling data heterogeneity,'' in \emph{CVPR}, 2024.

\bibitem{shi2024dataset_arxiv2024}
X.~Shi, W.~Zhang, M.~Wu, G.~Liu, Z.~Wen, S.~He, T.~Shah, and R.~Ranjan, ``Dataset distillation-based hybrid federated learning on non-iid data,'' \emph{arXiv preprint arXiv:2409.17517}, 2024.

\bibitem{jia2023unlocking_arxiv2023}
Y.~Jia, S.~Vahidian, J.~Sun, J.~Zhang, V.~Kungurtsev, N.~Z. Gong, and Y.~Chen, ``Unlocking the potential of federated learning: The symphony of dataset distillation via deep generative latents,'' in \emph{ECCV}, 2024.

\bibitem{dhasade2023quickdrop_arxiv2023}
A.~Dhasade, Y.~Ding, S.~Guo, A.-m. Kermarrec, M.~De~Vos, and L.~Wu, ``Quickdrop: Efficient federated unlearning by integrated dataset distillation,'' in \emph{Middleware}, 2024.

\bibitem{xu2024flip_arxiv2024}
S.~Xu, X.~Ke, X.~Su, S.~Li, F.~Xu, S.~Zhong \emph{et~al.}, ``Flip: Privacy-preserving federated learning based on the principle of least privileg,'' \emph{arXiv preprint arXiv:2410.19548}, 2024.

\bibitem{jin2025fedwsiddfederatedslideimage_arxiv2025}
H.~Jin, S.~Liu, C.~Cong, Q.~Feng, Y.~Liu, L.~Huang, and Y.~Hu, ``Fedwsidd: Federated whole slide image classification via dataset distillation,'' \emph{arXiv preprint arXiv:2506.15365}, 2025.

\bibitem{huang2024overcoming_icml2024}
C.-Y. Huang, K.~Srinivas, X.~Zhang, and X.~Li, ``Overcoming data and model heterogeneities in decentralized federated learning via synthetic anchors,'' in \emph{ICML}, 2024.

\bibitem{rub2024continual_icps}
M.~R{\"u}b, P.~Tuchel, A.~Sikora, and D.~Mueller-Gritschneder, ``A continual and incremental learning approach for tinyml on-device training using dataset distillation and model size adaption,'' in \emph{ICPS}, 2024.

\bibitem{lee2024practical_arxiv2024}
H.~Lee, J.~Lee, and N.~Kwak, ``Practical dataset distillation based on deep support vectors,'' \emph{arXiv preprint arXiv:2405.00348}, 2024.

\bibitem{chen2024distributed_cikm2024}
X.~Chen, W.~Meng, P.~Wang, and Q.~Zhou, ``Distributed boosting: An enhancing method on dataset distillation,'' in \emph{CIKM}, 2024.

\bibitem{wei2024physical_tpami2024}
H.~Wei, H.~Tang, X.~Jia, Z.~Wang, H.~Yu, Z.~Li, S.~Satoh, L.~Van~Gool, and Z.~Wang, ``Physical adversarial attack meets computer vision: A decade survey,'' \emph{IEEE Transactions on Pattern Analysis and Machine Intelligence}, 2024.

\bibitem{li2022backdoor_tnnls2022}
Y.~Li, Y.~Jiang, Z.~Li, and S.-T. Xia, ``Backdoor learning: A survey,'' \emph{IEEE Transactions on Neural Networks and Learning Systems}, 2022.

\bibitem{wu2024dd_arxiv2024}
Y.~Wu, J.~Du, P.~Liu, Y.~Lin, W.~Xu, and W.~Cheng, ``Dd-robustbench: An adversarial robustness benchmark for dataset distillation,'' \emph{arXiv preprint arXiv:2403.13322}, 2024.

\bibitem{zhou2024beard_arxiv2024}
Z.~Zhou, W.~Feng, S.~Lyu, G.~Cheng, X.~Huang, and Q.~Zhao, ``Beard: Benchmarking the adversarial robustness for dataset distillation,'' in \emph{CVPR}, 2025.

\bibitem{xue2024towards_arxiv2024}
E.~Xue, Y.~Li, H.~Liu, Y.~Shen, and H.~Wang, ``Towards adversarially robust dataset distillation by curvature regularization,'' in \emph{AAAI}, 2025.

\bibitem{liu2023backdoor_ndss}
Y.~Liu, Z.~Li, M.~Backes, Y.~Shen, and Y.~Zhang, ``Backdoor attacks against dataset distillation,'' in \emph{NDSS}, 2023.

\bibitem{chung2023rethinking_arxiv2023}
M.-Y. Chung, S.-Y. Chou, C.-M. Yu, P.-Y. Chen, S.-Y. Kuo, and T.-Y. Ho, ``Rethinking backdoor attacks on dataset distillation: A kernel method perspective,'' in \emph{ICLR}, 2024.

\bibitem{wu2024backdoor_arxiv2024}
J.~Wu, N.~Lu, Z.~Dai, W.~Fan, S.~Liu, Q.~Li, and K.~Tang, ``Backdoor graph condensation,'' \emph{arXiv preprint arXiv:2407.11025}, 2024.

\bibitem{chen2023comprehensive_arxiv2023}
Z.~Chen, J.~Geng, D.~Zhu, H.~Woisetschlaeger, Q.~Li, S.~Schimmler, R.~Mayer, and C.~Rong, ``A comprehensive study on dataset distillation: Performance, privacy, robustness and fairness,'' \emph{arXiv preprint arXiv:2305.03355}, 2023.

\bibitem{hu2022membership}
H.~Hu, Z.~Salcic, L.~Sun, G.~Dobbie, P.~S. Yu, and X.~Zhang, ``Membership inference attacks on machine learning: A survey,'' \emph{ACM Computing Surveys (CSUR)}, 2022.

\bibitem{moon2024towards_eccv2024}
J.-Y. Moon, J.~U. Kim, and G.-M. Park, ``Towards model-agnostic dataset condensation by heterogeneous models,'' in \emph{ECCV}, 2024.

\bibitem{zhao2023boosting_arxiv2023}
L.~Zhao, Y.~Zhang, F.~Chao, and R.~Ji, ``Boosting the cross-architecture generalization of dataset distillation through an empirical study,'' \emph{arXiv preprint arXiv:2312.05598}, 2023.

\bibitem{zhao2024metadd_arxiv2024}
Y.~Zhao, X.~Deng, X.~Su, H.~Xu, X.~Li, Y.~Liu, and S.~You, ``Metadd: Boosting dataset distillation with neural network architecture-invariant generalization,'' \emph{arXiv preprint arXiv:2410.05103}, 2024.

\bibitem{zhou2024improve_arxiv2024}
B.~Zhou, L.~Zhong, and W.~Chen, ``Improve cross-architecture generalization on dataset distillation,'' \emph{arXiv preprint arXiv:2402.13007}, 2024.

\bibitem{zhong2023towards_arxiv2023}
X.~Zhong and C.~Liu, ``Towards mitigating architecture overfitting in dataset distillation,'' \emph{arXiv preprint arXiv:2309.04195}, 2023.

\bibitem{han2022survey_tpami2022}
K.~Han, Y.~Wang, H.~Chen, X.~Chen, J.~Guo, Z.~Liu, Y.~Tang, A.~Xiao, C.~Xu, Y.~Xu \emph{et~al.}, ``A survey on vision transformer,'' \emph{IEEE transactions on Pattern Analysis and Machine Intelligence}, 2022.

\bibitem{wang2024dancing_cvpr2024}
Z.~Wang, Y.~Xu, C.~Lu, and Y.-L. Li, ``Dancing with still images: Video distillation via static-dynamic disentanglement,'' in \emph{CVPR}, 2024.

\bibitem{Chen2024ALS_arxiv2024}
Y.~Chen, S.-J. Guo, and L.~Wang, ``A large-scale study on video action dataset condensation,'' \emph{arXiv preprint arXiv:2412.21197}, 2024.

\bibitem{li2025latent_cvprw2025}
N.~Li, A.~A. Liu, J.~Zhang, and J.~Cui, ``Latent video dataset distillation,'' in \emph{CVPR Workshop}, 2025.

\bibitem{li2025videodatasetcondensationdiffusion_arxiv2025}
Z.~Li, H.~Reynaud, M.~Dombrowski, S.~Cechnicka, F.~X. Erick, and B.~Kainz, ``Video dataset condensation with diffusion models,'' \emph{arXiv preprint arXiv:2505.06670}, 2025.

\bibitem{ma2024latte_arxiv2024}
X.~Ma, Y.~Wang, G.~Jia, X.~Chen, Z.~Liu, Y.-F. Li, C.~Chen, and Y.~Qiao, ``Latte: Latent diffusion transformer for video generation,'' \emph{arXiv preprint arXiv:2401.03048}, 2024.

\bibitem{jiang2024ddfad}
W.~Jiang, R.~Zhang, H.~Li, X.~Liu, H.~Yang, and S.~Yu, ``Ddfad: Dataset distillation framework for audio data,'' \emph{arXiv preprint arXiv:2407.10446}, 2024.

\bibitem{zhang2025td3_www2025}
J.~Zhang, M.~Yin, H.~Wang, Y.~Li, Y.~Ye, X.~Lou, J.~Du, and E.~Chen, ``Td3: Tucker decomposition based dataset distillation method for sequential recommendation,'' in \emph{WWW}, 2025.

\bibitem{wu2024visionlanguage}
\BIBentryALTinterwordspacing
X.~Wu, B.~Zhang, Z.~Deng, and O.~Russakovsky, ``Vision-language dataset distillation,'' 2024. [Online]. Available: \url{https://openreview.net/forum?id=2y8XnaIiB8}
\BIBentrySTDinterwordspacing

\bibitem{xu2024low_icml2024}
Y.~Xu, Z.~Lin, Y.~Qiu, C.~Lu, and Y.-L. Li, ``Low-rank similarity mining for multimodal dataset distillation,'' in \emph{ICML}, 2024.

\bibitem{kushwaha2024audiovisual_tmlr2024}
S.~S. Kushwaha, S.~S.~N. Vasireddy, K.~Wang, and Y.~Tian, ``Audio-visual dataset distillation,'' \emph{Transactions on Machine Learning Research}, 2024.

\bibitem{hu2021lora_arxiv2021}
E.~J. Hu, Y.~Shen, P.~Wallis, Z.~Allen-Zhu, Y.~Li, S.~Wang, L.~Wang, and W.~Chen, ``Lora: Low-rank adaptation of large language models,'' in \emph{ICLR}, 2022.

\bibitem{zhang2025modalitycollapserepresentationsblending_arxiv2025}
X.~Zhang, Z.~Zhang, J.~Du, Z.~Liu, and J.~T. Zhou, ``Beyond modality collapse: Representations blending for multimodal dataset distillation,'' \emph{arXiv preprint arXiv:2505.14705}, 2025.

\bibitem{li2024dataset_arxiv2024}
M.~Li, C.~Cui, Q.~Liu, R.~Deng, T.~Yao, M.~Lionts, and Y.~Huo, ``Dataset distillation in medical imaging: A feasibility study,'' \emph{arXiv preprint arXiv:2407.14429}, 2024.

\bibitem{yu2024progressive_arxiv2024}
Z.~Yu, Y.~Liu, and Q.~Chen, ``Progressive trajectory matching for medical dataset distillation,'' \emph{arXiv preprint arXiv:2403.13469}, 2024.

\bibitem{smola2006maximum}
A.~J. Smola, A.~Gretton, and K.~Borgwardt, ``Maximum mean discrepancy,'' in \emph{ICONIP}, 2006.

\bibitem{li2024image_miccai2024}
Z.~Li and B.~Kainz, ``Image distillation for safe data sharing in histopathology,'' in \emph{MICCAI}, 2024.

\bibitem{guan2023discovering_arxiv2023}
H.~Guan, X.~Zhao, Z.~Wang, Z.~Li, and J.~Kempe, ``Discovering galaxy features via dataset distillation,'' in \emph{NeurIPS Workshop}, 2023.

\bibitem{willett2013galaxy}
K.~W. Willett, C.~J. Lintott, S.~P. Bamford, K.~L. Masters, B.~D. Simmons, K.~R. Casteels, E.~M. Edmondson, L.~F. Fortson, S.~Kaviraj, W.~C. Keel \emph{et~al.}, ``Galaxy zoo 2: detailed morphological classifications for 304 122 galaxies from the sloan digital sky survey,'' \emph{Monthly Notices of the Royal Astronomical Society}, 2013.

\bibitem{qifetch_nips2024}
D.~Qi, J.~Li, J.~Peng, B.~Zhao, S.~Dou, J.~Li, J.~Zhang, Y.~Wang, C.~Wang, and C.~Zhao, ``Fetch and forge: Efficient dataset condensation for object detection,'' in \emph{NeurIPS}, 2024.

\bibitem{hoiem2009pascal}
D.~Hoiem, S.~K. Divvala, and J.~H. Hays, ``Pascal voc 2008 challenge,'' \emph{World Literature Today}, 2009.

\bibitem{lin2014microsoft_eccv2014}
T.-Y. Lin, M.~Maire, S.~Belongie, J.~Hays, P.~Perona, D.~Ramanan, P.~Doll{\'a}r, and C.~L. Zitnick, ``Microsoft coco: Common objects in context,'' in \emph{ECCV}, 2014.

\bibitem{Dietz2025ASI_arxiv2025}
\BIBentryALTinterwordspacing
T.~Dietz, B.~B. Moser, T.~C. Nauen, F.~Raue, S.~Frolov, and A.~Dengel, ``A study in dataset distillation for image super-resolution,'' 2025. [Online]. Available: \url{https://api.semanticscholar.org/CorpusID:276161504}
\BIBentrySTDinterwordspacing

\bibitem{peng2025instancedatacondensationimage_arxiv2025}
T.~Peng, H.~M. Kwan, Y.~Jiang, G.~Gao, F.~Zhang, X.~Xu, S.~Liu, and D.~Bull, ``Instance data condensation for image super-resolution,'' \emph{arXiv preprint arXiv:2505.21099}, 2025.

\bibitem{zheng2025distributionawaredatasetdistillationefficient_arxiv2025}
Z.~Zheng, X.~Su, C.~Wu, and X.~Jia, ``Distribution-aware dataset distillation for efficient image restoration,'' \emph{arXiv preprint arXiv:2504.14826}, 2025.

\bibitem{cheng2024dataset_arxiv2024}
L.~Cheng, K.~Chen, J.~Li, S.~Tang, S.~Zhang, and M.~Wang, ``Dataset distillers are good label denoisers in the wild,'' \emph{arXiv preprint arXiv:2411.11924}, 2024.

\bibitem{wu2025trustawarediversiondataeffectivedistillation_arxiv2025}
Z.~Wu, Y.~Liu, X.~Shen, X.~Cao, and X.~Yu, ``Trust-aware diversion for data-effective distillation,'' \emph{arXiv preprint arXiv:2502.05027}, 2025.

\bibitem{lu2025unidetox_iclr2025}
H.~Lu, M.~Isonuma, J.~Mori, and I.~Sakata, ``Unidetox: Universal detoxification of large language models via dataset distillation,'' in \emph{ICLR}, 2025.

\bibitem{Qi_2025_CVPR_2025}
D.~Qi, J.~Li, J.~Gao, S.~Dou, Y.~Tai, J.~Hu, B.~Zhao, Y.~Wang, C.~Wang, and C.~Zhao, ``Towards universal dataset distillation via task-driven diffusion,'' in \emph{CVPR}, 2025.

\bibitem{jiang2024ddfad_arxiv2024}
W.~Jiang, R.~Zhang, H.~Li, X.~Liu, H.~Yang, and S.~Yu, ``Ddfad: Dataset distillation framework for audio data,'' \emph{arXiv preprint arXiv:2407.10446}, 2024.

\bibitem{yim2025permutation_arxiv2025}
J.-Y. Yim, D.~Kim, and J.-Y. Sim, ``Permutation-invariant and orientation-aware dataset distillation for 3d point clouds,'' \emph{arXiv preprint arXiv:2503.22154}, 2025.

\bibitem{xu2024low_arxiv2024}
Y.~Xu, Z.~Lin, Y.~Qiu, C.~Lu, and Y.-L. Li, ``Low-rank similarity mining for multimodal dataset distillation,'' \emph{arXiv preprint arXiv:2406.03793}, 2024.

\bibitem{sun2017revisiting_iccv017}
C.~Sun, A.~Shrivastava, S.~Singh, and A.~Gupta, ``Revisiting unreasonable effectiveness of data in deep learning era,'' in \emph{ICCV}, 2017.

\bibitem{schuhmann2022laion_nips2022}
C.~Schuhmann, R.~Beaumont, R.~Vencu, C.~Gordon, R.~Wightman, M.~Cherti, T.~Coombes, A.~Katta, C.~Mullis, M.~Wortsman \emph{et~al.}, ``Laion-5b: An open large-scale dataset for training next generation image-text models,'' in \emph{NeurIPS}, 2022.

\bibitem{li2025leveragingmultimodalinformationenhance_arxiv2025}
Z.~Li, H.~Reynaud, and B.~Kainz, ``Leveraging multi-modal information to enhance dataset distillation,'' \emph{arXiv preprint arXiv:2505.08605}, 2025.

\bibitem{li2024ddranking}
\BIBentryALTinterwordspacing
Z.~Li, X.~Zhong, Z.~Liang, Y.~Zhou, M.~Shi, Z.~Wang, W.~Zhao, X.~Zhao, H.~Wang, Z.~Qin, D.~Liu, K.~Zhang, T.~Zhou, Z.~Zhu, K.~Wang, G.~Li, J.~Zhang, J.~Liu, Y.~Huang, L.~Lyu, J.~Lv, Y.~Jin, Z.~Akata, J.~Gu, R.~Vedantam, M.~Shou, Z.~Deng, Y.~Yan, Y.~Shang, G.~Cazenavette, X.~Wu, J.~Cui, T.~Chen, A.~Yao, M.~Kellis, K.~N. Plataniotis, B.~Zhao, Z.~Wang, Y.~You, and K.~Wang, ``Dd-ranking: Rethinking the evaluation of dataset distillation,'' GitHub repository, 2024. [Online]. Available: \url{https://github.com/NUS-HPC-AI-Lab/DD-Ranking}
\BIBentrySTDinterwordspacing

\bibitem{valiant1984theory}
L.~G. Valiant, ``A theory of the learnable,'' \emph{Communications of the ACM}, 1984.

\bibitem{10740308_tpami2025}
E.~F. Montesuma, F.~M.~N. Mboula, and A.~Souloumiac, ``Recent advances in optimal transport for machine learning,'' \emph{IEEE Transactions on Pattern Analysis and Machine Intelligence}, 2025.

\bibitem{li2022awesome}
G.~Li, B.~Zhao, and T.~Wang, ``Awesome dataset distillation,'' \url{https://github.com/Guang000/Awesome-Dataset-Distillation}, 2022.

\end{thebibliography}

\end{document}